\documentclass[smallcondensed]{svjour3}  

\smartqed  

\usepackage{units}

\usepackage{bm}
\usepackage{float}
\usepackage[font=small,skip=7pt]{caption}
\usepackage[ruled,vlined,noend]{algorithm2e}
\usepackage{hyperref}
\usepackage{microtype}
\usepackage{multirow}
\usepackage{graphicx}
\usepackage[table,xcdraw]{xcolor}
\usepackage{epsfig}
\usepackage{rotating}
\usepackage[ruled,vlined]{algorithm2e}
\SetKwInput{KwInput}{Input}                
\SetKwInput{KwOutput}{Output}              
\newlength\fwidth

\usepackage{mathtools}

\usepackage{xcolor}
\usepackage{siunitx}
\usepackage{multirow}
\usepackage{epstopdf}
\usepackage{amsmath} 
\usepackage{amssymb}  
\usepackage{mathtools}
\usepackage[T1]{fontenc}
\DeclareGraphicsExtensions{.pdf,.eps}

\usepackage[acronym,nomain,nonumberlist]{glossaries}
\newacronym{mav}{MAV}{Micro Aerial Vehicle}
\newacronym{nmpc}{NMPC}{Nonlinear Model Predictive Control}
\newacronym{mpc}{MPC}{Model Predictive Control}
\newacronym{panoc}{PANOC}{Proximal Averaged Newton-type method for Optimal Control}
\newacronym{gps}{GPS}{Global Positioning System}
\newacronym{cnn}{CNN}{Convolutional Neural Network}
\newacronym{uwb}{UWB}{Ultra-Wide Band}
\newacronym{mbp}{MBP}{Motion Primitives-Based Path Planner}
\newacronym{compra}{COMPRA}{Compact Reactive Autonomy}
\newacronym{sar}{SAR}{Search-and-Rescue}
\newacronym{apf}{APF}{Artificial Potential Field}
\newacronym{dphr}{DPHR}{Deepest-point Heading Regulation}

\usepackage{url}

\begin{document}

\title{\LARGE \bf COMPRA: A COMPact Reactive Autonomy framework for subterranean MAV based search-and-rescue operations}
\titlerunning{COMPRA SubT exploration framework}        

  

\author{Bj\"orn Lindqvist$^1$ \and Christoforos Kanellakis$^1$ \and Sina Sharif Mansouri$^1$ \and Ali-akbar Agha-mohammadi$^2$ \and George Nikolakopoulos$^1$
}


\institute{Bj\"orn Lindqvist \at
              \email{bjorn.lindqvist@ltu.se}           
           \and
          Christoforos Kanellakis \at
              \email{christoforos.kanellakis@ltu.se}           
           \and
           Sina Sharif Mansouri \at
              \email{sina.sharif.mansouri@ltu.se}           
           \and
           Ali-akbar Agha-mohammadi \at
              \email{aliakbar.aghamohammadi@jpl.nasa.gov}           
           \and
           George Nikolakopoulos \at
              \email{george.nikolakopoulos@ltu.se}           
\and
$^1$ The authors are with the Robotics and Artificial Intelligence Team, Department of Computer, Electrical and Space Engineering, Lule\r{a} University of Technology, Lule\r{a} SE-97187, Sweden.
\and
$^{2}$The author is with Jet Propulsion Laboratory California Institute of Technology Pasadena, CA, 91109.
}

\date{Received: 2022-01-28 / Accepted: date}

\captionsetup{font=footnotesize}
\maketitle
\begin{abstract}\label{sec:abstract}
This work establishes COMPRA, a compact and reactive autonomy framework for fast deployment of \glspl{mav} in subterranean \gls{sar} missions. A COMPRA-enabled \gls{mav} is able to autonomously explore previously unknown areas while specific mission criteria are considered e.g. an object of interest is identified and localized, the remaining useful battery life, the overall desired exploration mission duration.
The proposed architecture follows a low-complexity algorithmic design to facilitate fully on-board computations, including nonlinear control, state-estimation, navigation, exploration behavior and object localization capabilities. The framework is mainly structured around a reactive local avoidance planner, based on enhanced Potential Field concepts and using instantaneous 3D pointclouds, as well as a computationally efficient heading regulation technique, based on depth images from an instantaneous camera stream. Those techniques decouple the collision-free path generation from the dependency of a global map and are capable of handling imprecise localization occasions.
Field experimental verification of the overall architecture is performed in relevant unknown \gls{gps}-denied environments.

\end{abstract}
\glsresetall 
%
\section{Introduction and Background}\label{sec:intro}
In recent years, aerial robotics research has experienced a rapid growth in different applications. Among the major objectives pursued, the overall aim is to increase the corresponding operating autonomy levels, envisioning the development and field deployment of aerial robotic workers capable to explore, inspect and operate in challenging environments without human intervention. In general, \glspl{mav} have undoubtedly shown powerful merits, as an outcome of their outstanding flying capabilities, in fully controlled and well-defined laboratory environments~\cite{kalantari2020drivocopter}, thus motivating the necessity to integrate such capabilities in real-life field applications. The envisioned use cases that are suitable for such aerial robotics operations, include disaster management missions~\cite{thakur2018nuclear}, infrastructure inspection and maintenance~\cite{mansouri2018cooperative}, subterranean area exploration~\cite{rogers2017distributed} and more. Focusing in \gls{sar} missions in subterranean environments, \glspl{mav} can provide access to unreachable, complex, dark and dangerous locations, while providing a valuable situational and environmental awareness to the first responders~\cite{subt}. Autonomous aerial robots can navigate in fully unknown and harsh environments and perform their instructed tasks, without the need of retaining a line of sight to the operator. In this approach, the human exposure to dangerous environments (e.g. blind openings, areas after blasting, etc.) will be significantly minimized, while at the same time will increase the effectiveness of the overall mission. Usually, in this type of environments, there is a lack of natural illumination, there exist unconditioned narrow passages and intersecting paths, dirt, high moisture and dust, which are factors making the environment harsh, complex and challenging for full autonomous missions.

The context of aerial robots in \gls{sar} missions, requires innovative and applicable in the field solutions that allow for a fast exploration of unknown areas, agile and collision free motion, as well as artifact detection and localization, all consisting of mission attributes that add resilience and robustness in these challenging missions. The focus of this work is the establishment of the COMPRA framework, especially designed for \glspl{mav}, including self-localization, \gls{nmpc}, obstacle avoidance, object localization and basic mission behaviour capabilities. A framework that has been extensively developed as part of the NEBULA autonomy~\cite{nebula,agha2021nebula,palieri2020locus,kim2021plgrim} for mission oriented drones in SubT environments and as an extension and differentiation from other co-existing complete frameworks for \gls{sar} operations.  
\subsection{Related Works}\label{sec:related_works}
 During the past years a significant amount of research works focused on the problem of robot path planning and navigation in 2D and 3D environments~\cite{mac2016heuristic,zhao2018survey,quan2020survey}. In general, the proposed frameworks support different type of environment representation, time complexity, as well as planning techniques (e.g. next-best-view, volumetric, random trees, graphs, etc ), which alter their performance in different types of environments. The recent DARPA SubT Challenge pushed the robotics community worldwide to develop the next generation of autonomous robotic explorers in SubTerranean environments to assist first responders in relevant \gls{sar} missions, providing them substantial situational awareness. Multiple research groups have participated in the SubT challenge and provided various methodologies, while the proposed COMRPA framework is our approach in the field of autonomous deployment of aerial robots underground. 
Within the related literature of full exploration mission frameworks in aerial robotics, there have been recently reported a number of works on methodologies for \gls{mav} deployment in subterranean environments.  

In~\cite{kratky2021autonomous} a framework for fast exploration in unknown 3D environments has been proposed, incorporating localization, mapping, planning, exploration and object detection and localization sub-components. The main focus of the system was on a grid-based path planner, enhanced with path post-processing, while the exploration behaviour was designed using a frontier-based strategy. The main sensing payload incorporated a 3D lidar and two RGB cameras. In \cite{ohradzansky2021multi} a vision-based local control architecture has been proposed for aerial vehicle SubT navigation. More specifically, a two-layers planning strategy has been developed, leveraging the map-based global path planning for exploration behaviour and an artificial potential-based approach that relied on depth information for local obstacle-free path following. The main sensing payload incorporated a 3D lidar and three RGB-D cameras.

In~\cite{dang2020autonomous} an autonomy framework for object search in underground areas has been proposed. The developed platform utilized onboard visual, thermal and 3D LiDAR sensors to address a multi-modal localization and mapping scheme. A bifurcated exploration planner is employed with local and global variants to explore the local surroundings of the platform, as well as to relocate the platform in globally defined unexplored paths or return to home. divided in local and global parts. The objects of interest are detected from a \gls{cnn} detector and localized using ray-casting in an occupancy map combined with binary Bayes filter to avoid erroneous measurements. Similarly, the authors in~\cite{petrlik2020robust} developed a \gls{mav} system for navigation in constrained tunnel-like environments. The main concept in this work, relies on a model based control scheme that relaxed the tracking performance to handle errors induced by the localization system. Moreover, an $A*$ planning approach on 2D occupancy maps is used for both global navigation and local exploration, while the objects of interest are detected using a \gls{cnn} and localized by an intersecting 3D rays from a camera projection model with wall and ground modeled by LiDAR sensors. The detected object can be reported to the DataBase either by following a return trajectory or through a designed mesh network when exploring the deepest parts.
In~\cite{sandino2020uav}, a framework for automated \gls{mav} motion planning under target location uncertainty in cluttered areas, has been presented in simulation and lab environments. In this work, the navigation problem is modelled as a Partially Observable Markov Decision Process (POMDP) and solved in real time through the Augmented Belief Trees (ABT) algorithm. Additionally, the developed system is able to handle a target detection uncertainty related to false positive detections by using the detection confidence in the POMDP formulation. In~\cite{ozaslan2017autonomous}, the estimation, navigation, mapping and control capabilities for autonomous inspection of penstocks and tunnels using aerial vehicles has been studied, using IMU, cameras and LiDAR sensors. In~\cite{mansouri2020deploying} the authors present an aerial scout robot for fast navigation in underground tunnels. The developed system enables a resource constrained robot to navigate as a floating object, combining a velocity controller on the $x$, $y$ and altitude control on the body frame axis, while proposing two different approaches for regulating the heading along the tunnel using either the geometric processing of 2D LiDAR scans or Deep Learning classification using monocular camera. \\
\vspace{-0cm}
Focusing in the qualitative comparison with the state-of-the-art works~\cite{dang2020autonomous}-\cite{mansouri2020deploying} the COMPRA architecture proposes an alternative strategy of the overall mission and the utilized modules related to localization, obstacle avoidance, exploration and object localization, while focusing in a low complexity implementation, as well as developing critical components to be resilient to  localization or occupancy mapping issues, including the reactive navigation based on Potential Fields obstacle avoidance, as well as a reactive heading regulation technique based on depth images. Moreover, the COMPRA core concept relies on the idea that smaller scaled \glspl{mav} for narrow passage navigation have limited flight time and in some application scenarios the higher priority is the fast deployment, compared to complete and detailed coverage of the environments. Table~\ref{tab:framework_comparison} summarizes the modules from all frameworks, depicting the alternative approaches to address the \gls{sar} mission.

\begin{sidewaystable}[!htpb]
\centering
\resizebox{\textwidth}{!}{%
\begin{tabular}{|c|c|c|c|c|c|c|c|}
\hline
\rowcolor[HTML]{FFFFFF} 
\textbf{Frameworks} & \textbf{Control} & \textbf{Localization} & \textbf{Obstacle Avoidance} & \textbf{Navigation} & \textbf{Return to Base} & \textbf{\begin{tabular}[c]{@{}c@{}}Object Detection \&\\ Localization\end{tabular}} & \textbf{Field Evaluation} \\ \hline
\rowcolor[HTML]{E6E6E6} 
\textbf{\cite{kratky2021autonomous,petrlik2020robust}} & \begin{tabular}[c]{@{}c@{}}Model Predictive Control\\ \& Acceleration tracking SO(3) controller \end{tabular} & 2D LiDAR odometry (Hector SLAM~\cite{kohlbrecher2011flexible}) & \begin{tabular}[c]{@{}c@{}}modified A* on\\ 2D occupancy grid\end{tabular} & \begin{tabular}[c]{@{}c@{}}modified A* on\\ 2D occupancy grid\end{tabular} & \begin{tabular}[c]{@{}c@{}}modified A* on\\ 2D occupancy grid\end{tabular} & Yolov3~\cite{redmon2016you} \& Raycasting & underground tunnel \\
\rowcolor[HTML]{FFFFFF} 
\textbf{\cite{ohradzansky2021multi}} & \begin{tabular}[c]{@{}c@{}}PX4 based flight controller\\ \& for attitude stabilization and velocity control \end{tabular} &  3D LiDAR SLAM (Google’s Cartographer~\cite{hess2016real}) & \begin{tabular}[c]{@{}c@{}}artificial potential field \\avoidance on depth images\end{tabular} & \begin{tabular}[c]{@{}c@{}}fast marching-based\\ frontier planning on graph\end{tabular} & \begin{tabular}[c]{@{}c@{}}graph-based\\ global planning\end{tabular} & Yolov3~\cite{redmon2016you} \& Depth Image & underground tunnel \\
\rowcolor[HTML]{E6E6E6} 
\textbf{\cite{dang2020autonomous}} & Model Predictive Control~\cite{kamel2017model} & \begin{tabular}[c]{@{}c@{}}Visual-Thermal-Inertial Odometry~\cite{khattak2020keyframe,dang2019field} \\ fused with LiDAR Odometry\end{tabular} & \begin{tabular}[c]{@{}c@{}}global-local \\ graph-based planners~\cite{dang2019graph} \\on 3D occupancy map\end{tabular} & \begin{tabular}[c]{@{}c@{}}local sampling-based\\rapidly exploring random\\ graph search planner\end{tabular} & \begin{tabular}[c]{@{}c@{}} global sampling-based\\rapidly exploring random\\ graph search planner\end{tabular} & Yolov3 \& Raycasting & underground tunnel \\
\rowcolor[HTML]{FFFFFF} 
\textbf{\cite{sandino2020uav}} & MAVROS position control~\cite{meier2015px4} & Realsense T265-Altimeter & occupancy map (offline) & \begin{tabular}[c]{@{}c@{}}POMDP decision making \\ using actions, states and observations \end{tabular}& - & \begin{tabular}[c]{@{}c@{}}MobileNet SSD\\ \& camera footprint with observation belief \end{tabular}& lab \\
\rowcolor[HTML]{E6E6E6} 
\textbf{\cite{ozaslan2017autonomous}} & \begin{tabular}[c]{@{}c@{}} Non Linear tracking control on SE(3)~\cite{lee2013nonlinear,mellinger2011minimum} \end{tabular} & \begin{tabular}[c]{@{}c@{}}range and vision-based odometry fused\\with Unscented Kalman Filter\end{tabular} & user high-level waypoint commands & \begin{tabular}[c]{@{}c@{}} user high-level waypoint commands\\ \& local mapper\\cylinder fitting \\parametric representation \\on 3D pointclouds \end{tabular} & - & - & underground tunnel \\
\rowcolor[HTML]{FFFFFF} 
\textbf{\cite{mansouri2020deploying}} & Velocity and altitude tracking \gls{nmpc} & OpticalFlow-Altimeter-IMU & 2D potential fields~\cite{kanellakis2018towards} & \begin{tabular}[c]{@{}c@{}}velocity based carrot chasing\\ \& tunnel axis detection \\ with either 2D LiDAR\\ or monocular CNN method\end{tabular} & \begin{tabular}[c]{@{}c@{}}tunnel axis yaw 180d rotation\\ \& velocity based carrot chasing\end{tabular} & - & underground tunnel \\
\rowcolor[HTML]{E6E6E6} 
\textbf{COMPRA} & Reference tracking \gls{nmpc}~\cite{lindqvist2020nonlinear,small2019aerial} & 3D LiDAR-Inertial odometry (LIO-SAM~\cite{shan2020lio}) & \begin{tabular}[c]{@{}c@{}}enhanced 3D \\ potential fields\\  \end{tabular} & \begin{tabular}[c]{@{}c@{}}Position based carrot chasing \\ \& depth-image open space\\  heading alignment\end{tabular} & \begin{tabular}[c]{@{}c@{}}breadcrumb following\\ on global waypoints\end{tabular} & Yolov4-tiny \& Depth Images & underground tunnel \\ \hline
\end{tabular}%
}
\caption{\label{tab:framework_comparison}\gls{mav} autonomy frameworks comparison}
\end{sidewaystable}

{The subterranean tunnel environment}\label{sec:environment}
The subterranean tunnel environment offers a set of specific challenges and opportunities when it comes to robot navigation. The physical environment itself of dark, dusty and narrow tunnels pose severe challenges to the utilized sensor configuration: completely dark areas limit the use of vision-based state estimation, and if RGB-cameras are to be used at all, the robot must carry its own illumination. Navigation (specifically obstacle avoidance) and state estimation pipelines have to be resilient to dust, where for example the UAV should avoid from flying too close to the ground to avoid excessive dust being disturbed from the propeller downwash. Self-similarity is also a major concern regarding state estimation and map-based loop-closure.
The combination of these factors with the narrow and curving tunnel areas can pose large problems to navigation. These tunnels heavily limit the field of view of onboard sensors, for example limiting the visual detection of ceiling and floor planes, and as such certain areas of the tunnel can be difficult to map completely, and it can be difficult to at all get some unknown/unmapped areas into the field of view of the robot without dangerous maneuvering (e.g. getting too close to walls, floor, ceiling). Obviously in narrow areas the margin of error for collision avoidance is heavily stressed as well as the robot is constantly very close to the walls. As such, it is to the authors experience and opinion from working in such environments, that classical map-based exploration architectures (especially in 3D) can struggle in those areas. Occupancy mapping, frontier generation, frontier selection, and occupancy-based path planning all rely on the ability to produce a consistent and, most importantly, a dense map of the environment, where a selected frontier point can be reached or seen without risking a crash, and where holes or unseen areas in the map don't promote unwanted or risky path planning behavior e.g. getting stuck, generating paths that go through holes or too close to the walls (a problem specifically analyzed in \cite{karlsson2021d}), or excessively moving back-and-forth in order to completely map the area due to hard-to-see frontiers. In general, for rapid deployment and efficient exploration, the proposed desired navigation behavior in these tunnels is to move in the center of the tunnel, at a good distance from the ground and ceiling, while following the tunnel direction and avoiding any encountered obstacles. And most importantly, to do so consistently, quickly, and with low rates of mission failure. The tunnel environment does present an opportunity as well, as the tunnel morphology itself can assist us in generating such behavior: by aligning the commanded locomotion of the \gls{mav} with the tunnel axis, while local collision avoidance can maintain a distance to walls and obstacles, the tunnel can be explored. It is towards solving these problems, while utilizing the tunnel morphology, that we propose a fully reactive navigation scheme that decouples the central navigation and exploration problem from the reliance on an accurate and dense global map, and guides the UAV through the tunnel environment by the simple motion directives of "follow the tunnel and avoid obstacles", which proved to be highly effective and consistent for navigating along narrow subterranean tunnels while also producing desirable navigation behavior in wider tunnels and voids.

\section{Target Mission Specifications}\label{sec:mission_spec}
While the COMPRA kit focuses on general autonomy requirements (nonlinear control, local reactive collision avoidance, exploration/navigation behavior, state estimation and object detection), the efficacy of the framework can be more easily evaluated for a specific mission. The targeted mission is in the \gls{sar} context, where rapid deployment is the major focus area. The COMPRA-enabled MAV should autonomously navigate through the subterranean tunnel environment, detect and localize objects of interest (for example survivors) and then return to base. Section \ref{sec:results} will focus on shorter directed missions (e.g. objects of interest are placed further into the unknown from the mission start position) in multiple different tunnel areas due to 1) limited battery life, 2) limitations on the size of testing environments where longer mission were not always possible due to impassable barriers or flight permissions, and 3) as COMPRA is a pure reactive scheme, it needs an addition of a higher-level mission planner to track junctions and to adapt the mission behaviour to explore more massive structures. Despite the third point, we shall show that COMPRA can on its own consistently perform the desired task of tunnel \gls{sar} under the targeted mission specifications, despite the difficulty of navigating the narrow tunnel subterranean terrain.

\subsection{Contributions}\label{sec:contributions}
With respect to the envisioned application environment and mission scenario, and the related State-of-the-Art, the main contributions of our work are listed as follows:
\begin{itemize}

    \item A complete autonomy framework that has all necessary components for full mission execution in the subterranean \gls{sar} context, including: sensor suite selection, state estimation, nonlinear control, obstacle avoidance, exploration, object detection and localization, and mission behavior to allow for complete mission execution. 
    
    \item A fully reactive navigation architecture that is based around an \gls{apf} formulation that requires no in-between software and can directly use the LiDAR 3D pointcloud stream for safe navigation, as well as as an effective heading-regulation technique that utilizes depth-images in order to attract the \gls{mav} towards open areas. The combination of \gls{apf} and heading regulation results in a fully reactive scheme for quick deployment and fast subterranean tunnel navigation, that completely decouples the navigation behavior from the reliance on a global map, as well as from occupancy-based path planning or frontier generation. The navigation framework has been developed specifically to allow stable, fast, and consistent navigation in narrow subterranean environments while remaining compact and low complexity.
    
    \item A complete object detection and localization pipeline with multiple layers of false-positive rejection, evaluated in completely dark areas using only onboard illumination, as well as in areas that specifically challenge detection of specific objects of interest.

    \item An extensive experimental validation in a variety of subterranean tunnel environments focusing on narrow tunnels, wide tunnels, voids, dust, as well as visits to real mining environments. The evaluation of the overall architecture focuses on full mission execution, with all components for realistic \gls{sar} in the loop. The emphasis is on quick deployment and rapid exploration for time-constrained missions. A complete comparisons with the identified similar mission frameworks cannot be performed due to the lack of complete software open source packages that would allow such action, however insights and highlights on the different point of view have been provided. We do provide a brief comparison on the problem of agile subterranean tunnel navigation in real-life field evaluations, comparing COMPRA to a state-of-the-art local exploration-planning framework.
  
\end{itemize}

The rest of the article is structured as follows. Initially, Section~\ref{sec:methodology} presents detailed information of the COMPRA autonomy stack, while discussing both the baseline navigation framework, as well as the capabilities related to the mission definition. The experimental evaluation of the framework and the corresponding results are presented in Section~\ref{sec:results}. Section \ref{sec:limitations} discuss discovered challenges and limitations to the framework, and offer directions for future work. Finally, Section~\ref{sec:conclusion} concludes the article by summarizing the findings.
\section{COMPRA Autonomy} \label{sec:methodology}

\subsection{Frame Notation}\label{sec:frames}
\begin{figure}[!htpb]
    \centering
    \includegraphics[width=0.8\textwidth]{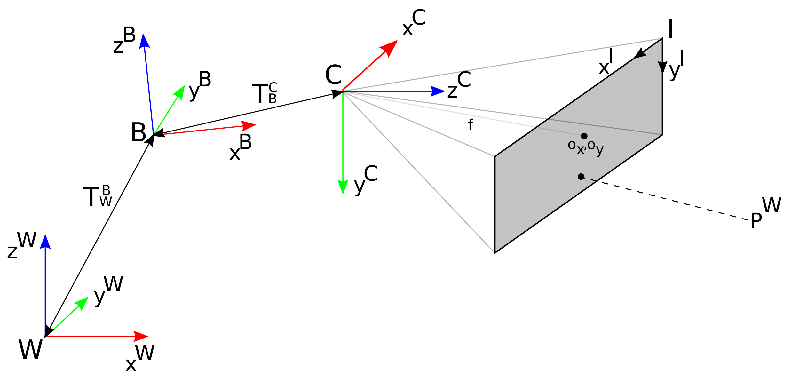}
    \caption{Coordinate frames, where $^\mathcal{W}$, $^\mathcal{B}$, $^\mathcal{L}$, $^\mathcal{C}$ and $^\mathcal{I}$ denote world, body, LiDAR, camera and image coordinate frames respectively.}
    \label{fig:coord_frames}
\end{figure}

The world frame $\mathcal{W}$ is fixed with the unit vectors $\{x^{\mathcal{W}}, y^{\mathcal{W}}, z^{\mathcal{W}}\}$ following the East-North-Up (ENU) frame convention. The body frame of the aerial vehicle $\mathcal{B}$ is attached on its base with the unit vectors $\{x^{\mathcal{B}}, y^{\mathcal{B}}, z^{\mathcal{B}} \}$, representing the rotated global coordinates $\mathcal{W}$ in  along the $z$-axis. The  $z^{\mathcal{B}}$ is antiparallel to the gravity vector, $x^{\mathcal{B}}$ is looking forward the platform's base and $y^{\mathcal{B}}$ is in the ENU convention. The onboard camera frame $\mathcal{C}$ has unit vectors $\{x^{\mathcal{C}}, y^{\mathcal{C}}, z^{\mathcal{C}} \}$. Furthermore, $y^{\mathcal{C}}$ is parallel to the gravity vector and $z^{\mathcal{C}}$ points in front of the camera. The image plane is defined as $\mathcal{I}$ with unit vectors $[x^\mathcal{I},y^\mathcal{I}]$. 
Figure~\ref{fig:coord_frames} depicts the utilized main coordinate frames of the aerial platform.

\subsection{Navigation Framework}\label{sec:autonomy}
\begin{figure}[!htpb]
    \centering
    \includegraphics[width=\textwidth]{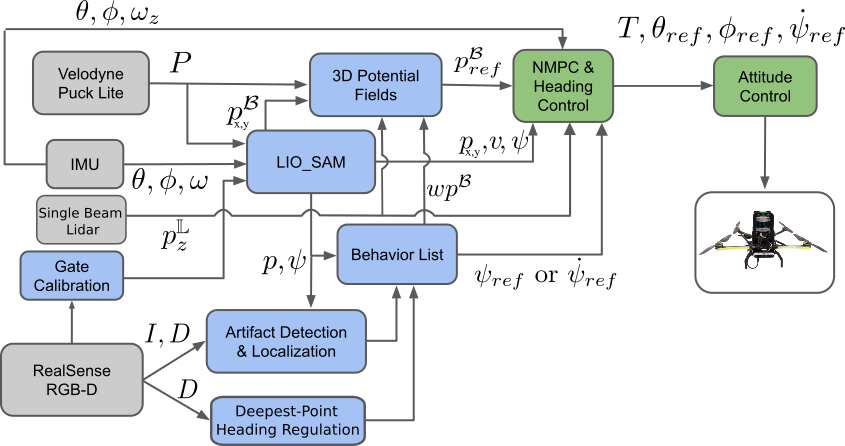}
    \caption{Overall COMPRA architecture and information flow. On-board sensors are shown in gray, sub-modules/algorithms in blue, and controllers in green.}
    \label{fig:architecture}
\end{figure}

In this part we present the fundamental systems to enable the autonomous flight of a \gls{mav} along tunnels by using onboard computation unit and sensors. The baseline functionality for the aerial platform is to accomplish an exploration task in unknown underground areas to localize objects of interest. To this end, the COMPRA framework incorporates state estimation, control, avoidance, exploration and object detection and localization capabilities. Initially, it localizes itself in a globally defined coordinate frame combining information of a fiducial based gate calibration scheme with the LiDAR-Inertial based state estimation just before the mission starts. The obstacle avoidance scheme is based on \glspl{apf} and generates collision-free waypoints. The reactive exploration behaviour guides the \gls{mav} along the tunneling environment, while, the control reference tracker uses the information from the other components to provide low level commands for the flight control unit. In the proposed system a single beam LiDAR facing towards the ground provides altitude measurements relative to the ground $p^\mathbb{L}_z$, which are used from other COMPRA subsystems. Figure~\ref{fig:architecture} summarizes COMPRA exploration framework.
\subsubsection{State-Estimation}\label{sec:localization}
The autonomous flight of a \gls{mav} requires state-estimation information that is accurate and with low-latency. Generally, research on robot localization has been and continues to be a quest with the aim to establish robotics into everyday life usage. Multiple sensing modules can be used to provide state-estimation information including monocular/depth/stereo/event/thermal cameras, 2D/3D LiDARs, Radars, \gls{gps}, \gls{uwb} and other, depending on the application needs. In this framework the State-of-the-Art tightly coupled 3D LiDAR-Inertial odometry estimation scheme LIO-SAM, proposed in~\cite{shan2020lio} is used that provides the resulting state vector $X=[p_x,p_y,p_z,v_x,v_y,v_z,\phi,\theta, \psi, \omega_x, \omega_y, \omega_z]^\top$. The COMPRA framework integrated this specific LiDAR-based SLAM approach, apart form being a SotA method , because it is structured in an efficient graph-based optimization framework and map management for high rate odometry estimation as well as supporting loop closing capabilities, while it can be extended with additional sensor inputs. Being LiDAR-based is not reliant on illumination factors, and fast IMU-based predictions facilitate the high run-time requirements of the \gls{mav} platform.
%
\normalsize 
\subsubsection{Reference Tracking Controller}\label{sec:mpc}
While there are a wide range of possible reference tracking controllers for \glspl{mav} we use a \gls{nmpc}. The main benefits of \gls{nmpc} are the ability to handle constraints, the flexibility in defining the cost function, and that it is directly based on a nonlinear dynamic model of the system. The proposed NMPC is similar in structure to previous papers \cite{lindqvist2020nonlinear}, \cite{small2019aerial}, and as such we will not go into details. The MAV nonlinear model considers eight states namely $x = [p,v,\phi, \theta]$, and as such it is in a yaw-compensated coordinate frame. The generated control commands are $u = [T, \theta_{ref}, \phi_{ref}]$ with $\phi_{\mathrm{ref}}\in \mathbb{R}$, $\theta_{\mathrm{ref}}\in \mathbb{R}$ and $T_{\mathrm{ref}}\geq 0$ to be the references in roll, pitch and total mass-less thrust generated by the four rotors, which are very commonly accepted commands, together with a yaw-rate command as $\Dot{\psi}_{ref}$, for low level attitude controllers, such as Pixhawk or ROSflight, with $T_{\mathrm{ref}}$ commonly mapped to a control signal as $u_t \in [0,1]$. The yaw angle $\psi$ is controlled with a decoupled simple PD controller.
Let $x_{k+j{}\mid{}k}$ denote the predicted state 
at time step $k+j$, produced at the time step $k$ (and similarly the control action as $u_{k+j{}\mid{}k}$), and $N$ the prediction horizon. We formulate the cost functions to make each state reach the prescribed set-points, while delivering smooth control signals as:
%
\begin{eqnarray}
\label{eq:costfunction}
J({x}_{k}, {u}_{k}; u_{k-1\mid k}) = \sum_{j=0}^{N}   \underbrace{\| x_{\mathrm{ref}}-x_{k+j{}\mid{}k}\|_{Q_x}^2}_\text{State cost} 
\\\nonumber
+ \underbrace{\| u_{\mathrm{ref}}-u_{k+j{}\mid{}k}\|^2_{Q_u}}_\text{Input cost}
+  \underbrace{\| u_{k+j{}\mid{}k}-u_{k+j-1{}\mid{}k} \|^2 _{Q_{\Delta u}}}_\text{Input change cost},
\end{eqnarray}
\normalsize
where $Q_x\in \mathbb{R}^{8\times8}, Q_u, Q_{\Delta u}\in 
\mathbb{R}^{3\times3}$ are symmetric positive definite weight matrices for the
states, inputs and input rates respectively. Importantly we can directly penalize the change in inputs from one time step to the next by the \textit{input change cost}, promoting non-aggressive control actions. Additionally let us define constraints on the inputs as $u_{\min} \leq u_{k+j\mid k} \leq u_{\max}$ and constraints on the change in control inputs, to further enforce smooth control actions, as:
\begin{subequations}
\begin{align}
    |\phi_{\mathrm{ref}, k+j-1{}\mid{}k} - \phi_{\mathrm{ref},k+j{}\mid{}k}| 
    {}\leq{}
    \Delta \phi_{\max},
    \\
    |\theta_{\mathrm{ref}, k+j-1{}\mid{}k} - \theta_{\mathrm{ref},k+j{}\mid{}k}| 
    {}\leq{}
    \Delta \theta_{\max}.
\end{align}
\end{subequations}
where $\Delta \phi_{\max}$ and $\Delta \theta_{\max}$ denote the maximum allowed change in roll and pitch references from one time step to the next. The optimization problem to minimize the cost function $J(\bm{x}_{k}, \bm{u}_{k}; u_{k-1\mid k})$, while subject to the constraints is solved by the Optimization Engine\cite{sopasakis2020open}, an open-source and Rust-based parametric optimization software, which is very fast for this type of application. While the input constraints are considered as hard bounds, OpEn utilizes a penalty method\cite{Hermans:IFAC:2018} to solve for trajectories that satisfy the input rate constraints. The NMPC runs at 20Hz, which is following common inner/outer loop dynamics with the attitude controller running at 100Hz. The prediction horizon $N$ that is considered in the optimization is 20, implying a one second prediction. 

\subsubsection{Obstacle Avoidance}\label{sec:avoidance}
There are a number of requirements that we pose on the local avoidance scheme, such as to: 1) be reactive e.g. have low computation and latency, 2) be independent of localization and mapping as to prevent crashes due to localization or occupancy-mapping failure, 3) result in smooth and stable behavior of the \gls{mav} in any avoidance situation, 4) work directly with the 3D point cloud (again to prevent failures due to drifts or malfunctions in other software). Towards these requirements we propose an \gls{apf} formulation with a focus placed on generating a resulting force in the local \gls{mav} frame that does not cause oscillations or twitching flight behavior of the \gls{mav}. This is achieved by choosing a potential function (or rather directly a force function) that is continuous and smooth in the area of influence of the potential field, placing saturation limits on the repulsive forces and the change in repulsive forces, and performing force-normalization, as to always generate forces of the same magnitude (which a reference-tracking controller then can be optimally tuned to follow). 

Let us denote the local point cloud generated by the 3D LiDAR as $\{P\}$, where each point is described by a relative position to the LiDAR frame as $\rho = [\rho_x, \rho_y, \rho_z]$. Let us also denote the repulsive force as $F^r = [F_{x}^r, F_{y}^r, F_{z}^r]$, the attractive force as $F^a = [F_{x}^a, F_{y}^a, F_{z}^a]$, the radius of influence of the potential field as $r_F$. As we are only interested in points inside the radius of influence, when considering the repulsive force, let's denote the list of such points $\boldsymbol{\rho}_r \in \{P\}$ where $\mid\mid \rho_r^i \mid \mid \leq r_F$ and $i = 0,1, \ldots, N_{\rho_r}$ (and as such $N_{\rho_r}$ is the number of points to be considered for the repulsive force). Similarly, we impose an inner safety-critical radius $r_c$ with a large static potential where $N_{\rho_c}$ is the number of points $\rho_c$ inside this radius. 
Taking inspiration from the classical repulsive force function proposed by Warren~\cite{warren1989global} we define the repulsive force as: 
\begin{equation}\label{eq:repulsive}
    F^r = \sum^{N_{\rho_r}}_{i=0}  L(1 -  \frac{\mid\mid \rho_r^i \mid\mid}{r_F})^2\frac{-\rho_r^i }{\mid\mid \rho_r^i \mid\mid} + \sum^{N_{\rho_c}}_{i=0}  L_c\frac{-\rho_c^i }{\mid\mid \rho_c^i \mid\mid}
\end{equation}
\normalsize
where $L = [L_x, L_y, L_z]$ is the repulsive constant and represents the largest possible force-per-point inside $r_F$ and $L_c$ represents the magnitude of the static potential to ensure no obstacle enters $r_c$. It should also be noted that computing the total force as the sum of multiple smaller "forces" from each point inside $r_F$ also makes the framework more resilient to dust as each individual dust particle does not majorly affect the total force. As the repulsive force from the inner radius necessarily imposes a larger force-per-point we shall impose that it is zero if $N_c \leq n$ with $n$ being a small number such that it is resistant to dust while still detecting small obstacles.

Based on the COMPRA framework, we simply denote the attractive force as the next way-point, $wp = [wp_x, wp_y, wp_z]$ generated by the navigation in the local frame, such that $F^a = wp^{\mathcal{B}} - \hat{p}^{\mathcal{B}}$. From an intuitive point of view this can be seen as the attractive force being the vector from the current position to the next given way-point with an unitary gain, while the repulsive force is the shift in the next way-point required to avoid obstacles.
Generally, the attractive and repulsive forces are summed to get the total resulting force $F$. But, since the requirement of a stable and smooth flight is very high for a \gls{mav} as with multiple other modules (localization etc.) running from on-board sensors, we propose a force-normalization and saturation on the forces, and on the change in the repulsive force. Let $k$ denote the current sampled time instant and $(k-1)$ the previous sampled time instant. This process is explained in Algorithm 1, with $F_{max}$ and $\Delta F_{max}$ being the saturation value and rate-saturation of the repulsive force and $sgn()$ denoting the sign function. The reference position, passed to the reference tracking controller (in the \gls{mav} body frame), with included obstacle avoidance, then becomes $p_{ref}^{\mathcal{B}} = F + \hat{p}^{\mathcal{B}}$, where $p_{ref}^{\mathcal{B}}$ are the first three elements of $x_{\mathrm{ref}}$. 

\begin{algorithm}[htbp]
\SetAlgoLined
\textbf{Inputs:} $F^a, F^r_k, F^r_{k-1}$  \\
\If{$\mid\mid F^r_k \mid\mid > F_{max}$}{ 
        $F^r_k \gets sgn(F^r_k)F_{max}$} 
\If{$\mid\mid F_k - F_{k-1}\mid\mid > \Delta F_{max}$}{ 
        $F^r_k \gets F^r_{k-1} + sgn(F_k^r - F^r_{k-1})\Delta F_{max}$ 
}
\If{$\mid\mid F^a \mid\mid > 1$}{
    $F^a \gets \frac{F^a}{\mid\mid F^a \mid\mid}$}
$F \gets F^r_k + F^a$ \\
\If{$\mid\mid F \mid\mid > 1$}{
    $F \gets \frac{F}{\mid\mid F \mid\mid}$}
\textbf{Output:} $F$  
\caption{Force calculation}\label{alg:force_calc}
\end{algorithm}

Additionally, we propose that the desired flight behavior when in the presence of obstacles is not the same as when moving in obstacle-free areas. As the MAV gets closer to obstacles, the local avoidance behavior should be to move slower and with less rapid maneuvering, e.g. being more careful. As such, we impose a simple adaptive weights scheme where the computed repulsive potential field forces are used to adapt the NMPC weight matrix $Q_x$, or more specifically the first three elements in $Q_x$ that considers the position states here denoted as $Q_p$, as to decrease the emphasis on position reference tracking with an increase in the force magnitude (before force normalization). We map $Q_p$ between minimum and maximum values $Q_{p,min}, Q_{p,max}$ as:
\begin{equation}
    Q_p = Q_{p,min} + \frac{Q_{p,max} - Q_{p,min}}{1 + c \mid\mid F_r \mid\mid},
\end{equation}
with $c$ being a tuning constant. Despite a relatively simple expression, this greatly assists the MAV in moving carefully when in-between obstacles or walls in narrow areas, while still moving as rapidly as possible when the area is obstacle free. This allows us to safely use a larger $Q_{p,max}$ (for faster reaching of way-points) without risking collisions and unwanted flight behavior when in the presence of obstacles or when entering a narrow area.

\subsubsection{Reactive Exploration}\label{sec:exploration_planner}

The exploration behaviour in this work considers the generation of local reference waypoints $p_{ref}^{\mathcal{B}}$ in 3D, as well as local yaw references $\psi_{ref}$.

The exploration waypoints follow the heuristic concept of constant value ``carrot'' chasing in the bodyframe $x$-axis of the platform, while we utilize a fully reactive heading regulation technique to align the MAV body \textit{x}-axis with the tunnel direction. More specifically, the generated way-points $p_{ref}^{\mathcal{B}}$ always add a constant value ahead of the $x$-axis, while the motion in the $y$-axis in $\mathcal{B}$ frame depends only on the potential fields input. The waypoints are defined as $wp^{\mathcal{B}} = [p_x^{\mathcal{B}}+1, p_y^{\mathcal{B}}, p_z^{m}]$, and are then fed to the potential field to generate the obstacle-free $p_{ref}^{\mathcal{B}}$. The local altitude reference $p_z^{m}$ (or the mission altitude) is kept constant and selected before the mission starts, and the measured local \textit{z}-position is defined by the range measurements, $R_{sbl}$, from the single-beam LiDAR as $p^\mathbb{L}_z = R_{sbl}\cos{\theta}\cos{\phi}$, e.g. the distance to the ground.

\begin{figure}[!htbp]
    \centering
\includegraphics[width=0.8\columnwidth]{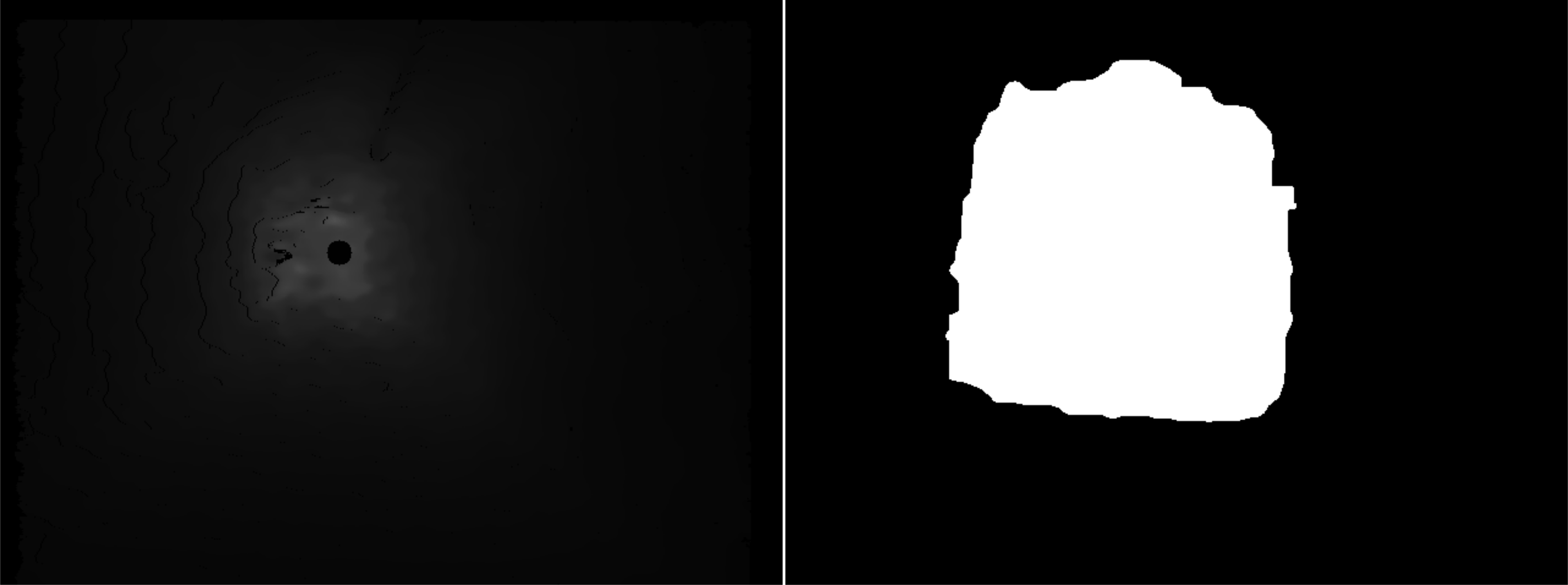}
    \caption{Snapshots of the \gls{dphr} methodology depicting, on the left the extracted centroid (marked black circle) of the open tunnel area ahead and on the right the binarized image of the segmented area with the higher depth values.}
    \label{fig:DPHR}
\end{figure}

\paragraph{Deepest-point Heading Regulation}\label{sec:heading}
Part of the fast deployment and exploration approach of COMPRA is the alignment of the aerial platform heading towards open/deep areas, resulting in the \gls{mav} being attracted to open areas and aligns itself to the tunnel direction.
The \gls{dphr} technique utilizes the onboard RGB-D instantaneous camera stream to reactively find the deepest cluster of points within the stream. Initially, the recovered depth images from the sensor are filtered using a grey scale morphological close operation~\cite{Soille_2003} as a preprocessing step to remove noise and enhance the tunnel opening. Afterwards, a clustering step is employed using a k-means methodology to extract a fixed number of clusters $C_i,[i=1,2,3, \dots, N_{clusters}]$, where the $N_{clusters}=10$ value has been selected based on the tunnel environment morphology. Moreover, the mean intensity value for each cluster region is calculated and we select the cluster with the maximum intensity, which indicates the deepest parts of the tunnel. The $x$-axis pixel coordinate of the cluster centroid is calculated as $s_x = \frac{1}{|C_m|}\sum_{(x,y) \in C_m}^{} x$, where $|C_m|$ represents the number of pixels. Finally, $s_x$ is normalized and transformed with respect to the image principal point $\bar{s_x}$ and converted to a yaw rate reference $\dot{\psi}_{ref}$ $\in$ [min max], using $\dot{\psi}_{ref}=\bar{s_x}*l$, where $l$ maps linearly the yaw rate to min and max values. Therefore, the yaw angle $\psi$, which controls the UAV body $x$-axis direction is aligned with the tunnel direction based on the heading regulation technique. The technique is fully reactive, low computation, and allows fast and consistent navigation in the subterranean areas. Figure~\ref{fig:DPHR} showcases a snapshot from the implemented deepest point extraction process.

\subsection{Mission-specific Modules}\label{sec:mission}
For the overall mission definition of \gls{sar} in SubT environments, two components must be added to the core COMPRA autonomy kit to fulfill the mission, namely: the object detection and localization based on an image stream from onboard cameras, as well as a behavior list for mission execution, that includes returning to base after mission completion.
\subsubsection{Object Detection and Localization}\label{sec:artifacts}

In this Section we consider the capability to detect and localize objects of interest in previously unknown environments, using the flying platform's onboard sensor data. This process relates to \gls{sar} missions, where in this work the object detector uses the visual spectrum (RGB images) and the object localizer uses depth image measurements.  The object detection part is based on the tiny and Intel hardware optimized version~\cite{openvino} of the state of the art CNN object detector Yolo V4~\cite{bochkovskiy2020yolov4}.
We trained the network to detect and classify 6 classes $CL\in[1,\cdots,6]$ defined by the SubT competition using a custom dataset consisting of approximately 700 images for each class. The input size of the images is 416 $\times$ 416 and the output of the algorithm are the detected bounding boxes and the class probability $Pr_{CL} \in [0,1]$. 

\begin{figure}[!htbp]
    \centering
\includegraphics[width=0.8\columnwidth]{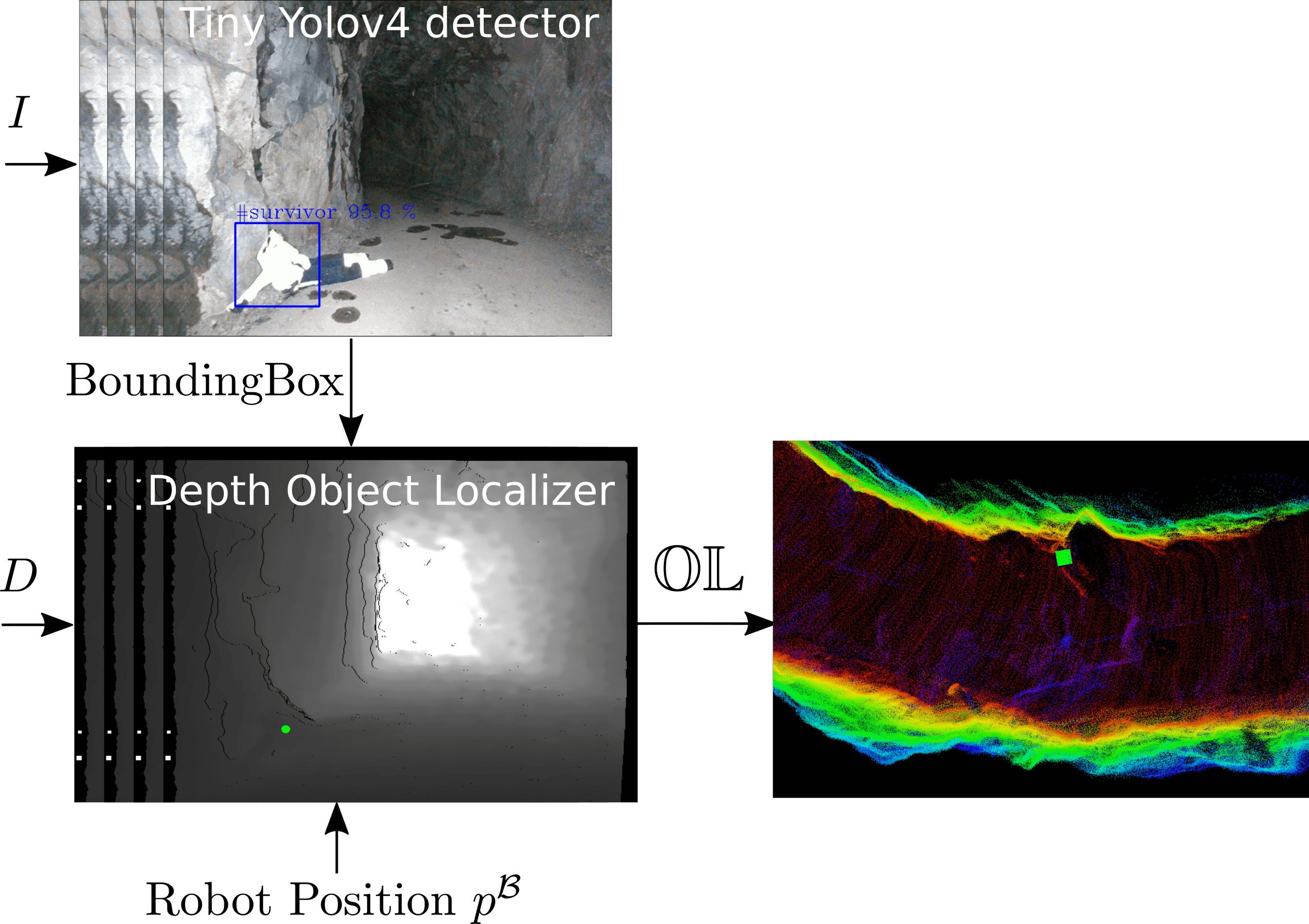}
    \caption{The object detection and localization pipeline.}
    \label{fig:obj_loc}
\end{figure}

The other component of the pipeline is the object localizer, which receives the bounding box $BB=(x_{min}^\mathcal{I}, y_{min}^\mathcal{I}, Wd^\mathcal{I}, Ht^\mathcal{I})$ measurements of one of the predefined object classes from the RGB image stream $I$, where $x_{min}^\mathcal{I}$ and $y_{min}^\mathcal{I}$ denote the minimum $x^\mathcal{I}$ and $y^\mathcal{I}$ axis pixel coordinates and $ Wd^\mathcal{I}, Ht^\mathcal{I}$ denote the width and height in pixels. The localizer transfers the identified bounding box in the aligned depth image stream $D$ and extracts the relative position of the object in the camera frame $\mathcal{C}$, defined as $p_{object}^{\mathcal{C}} = [p_x^{\mathcal{C}}, p_y^{\mathcal{C}}, p_z^{\mathcal{C}}]$. Frequently, the extracted bounding boxes include part of the background with the object of interest, thus to avoid this issue we calculate the object position considering only a 3$\times$3 window around the centroid of the bounding box. The main assumption is that the centroid always is projected to the object of interest.
Finally the object location is converted in the global world frame $\mathcal{W}$, defined as $p_{object}^{\mathcal{W}}$, using the transformation $p_{object}^{\mathcal{W}} = {}^{\mathcal{W}}T_{\mathcal{C}} ~p_{object}^{\mathcal{C}}$, where ${}^{\mathcal{W}}T_{\mathcal{C}}$ denotes the transformation matrix from $\mathcal{C}$ to $\mathcal{W}$, defined as ${}^{\mathcal{W}}T_{\mathcal{C}}=[R|t]$. The object localizer is structured around two subcomponents, i) the buffer of measurements and ii) the processor of buffered measurements. The buffer stacks artifact positions using a two-step outlier rejection process. Initially, it accepts only bounding boxes with class probability above a specified threshold ($Pr_{CL} \in [Pr_{threshold},1]$) and are located inside a sphere with radius of 5 meters and secondly, removes detections when their metric bounding box width bounds are outside a fixed width interval for each known object $ Wd^\mathcal{W} \in [Width^{CL}_{min},Width^{CL}_{max}]$, to address false positive inputs with high class probabilities. Afterwards, the processor of buffered measurements is initiated once a specified time window from the last observation in the buffer has passed. During this process, the buffered values for each class are clustered based on Euclidean distance and the mean value of the positions of each cluster is calculated. Additionally, the current clusters are compared against already localized objects using Euclidean distance to deduce whether it will belongs to previously seen object or it will be reported as a new observation. Once this step is finished the buffer is cleaned. This architecture can handle multiple observations of the same class at different locations in the same buffer session. 
The localizer returns a list  $\mathbb{OL}=\{\vec{0}_{3\times1}, p_{object1}^{\mathcal{W}}, p_{object2}^{\mathcal{W}}, \dots, p_{objectn}^{\mathcal{W}}\}$, where $n$ is the number of detected objects with $n\in \mathbb{N}_0$.  Figure~\ref{fig:obj_loc} presents the overall architecture on the strategy related to the object detection and localization.
\subsubsection{Behavior List}\label{sec:tree}

The mission behavior of COMPRA is presented in Algorithm~\ref{alg:behavior}

\begin{algorithm}[!htbp]
\DontPrintSemicolon
  
  \KwInput{Mission Plan}
  \KwOutput{Object List $\mathbb{OL}=\{\vec{0}_{3\times1}, p_{object1}^{\mathcal{W}}, p_{object2}^{\mathcal{W}}, \dots, p_{objectn}^{\mathcal{W}}\}$}
  Gate Calibration \tcp*{{$\mathcal{W}$} frame definition}
  Pre-Flight checks \tcp*{software and hardware ready to fly \gls{mav}}
  
  \If{Pre-Flight checks}
    {
        Take-off Sequence \tcp*{stabilize hover at $p_z^{m}$}
        \If{Take-off Sequence}
        {
        Exploration=True \tcp*{start~exploration}
        \While{Exploration}
            {
            \If{object detected}
            {
                \While{object in view at time instance K}
                    {
                        $\mathbb{P}_{object}^{\mathcal{C}}=[\mathbb{P}_{object}^{\mathcal{C}};p_{object,K}^{\mathcal{C}}]$ 
                    }
                \If{size($\mathbb{P}_{object}^{\mathcal{C}}$)$>$0}
                {
                objectLocalized=True
                }
            }
                
            \If{objectLocalized}
            {
            $p_{object}^{\mathcal{W}}=\overline{\mathbb{P}}_{object}^{\mathcal{C}}$ \\
            $\mathbb{OL}=\{\mathbb{OL}; p_{object}^{\mathcal{W}}\}$\\
            objectLocalized=False
            }
                
                \If{$T^{current}_{mission} \leq T^{reference}_{mission}$  }
                {
                $\Dot{\psi}_{ref}$, $wp^{\mathcal{B}} = [p_x^{\mathcal{B}}+1, p_y^{\mathcal{B}}, p_z^{m}]$
                }
                Breadcrumb Waypoint List $\mathbb{BW}= [\mathbb{BW}{;}~wp^{\mathcal{B}} \psi]$
                
                \Else
                {
                    $\psi^{ref}_{return} = \psi + \pi$ 
                    
                    size($\mathbb{BW}$)$=k$
                    
                    $wp^{\mathcal{B}} = \mathbb{BW}(k)$ 
                    
                    $\mathbb{BW} \symbol{92} \mathbb{BW}(k) $ \tcp*{remove $k$th entry from  $\mathbb{BW}$, $ \symbol{92}$ denotes relative complement operation  }
                    
                    \If{$\mathbb{BW}=\{ \emptyset \}$}
                    {
                        Exploration=False
                        
                        Land \tcp*{mission completed}
                    }
                }
            }
            
        }
    }
    \Else
    {
        Hardware and Software health check
    }
 \caption{COMPRA mission workflow}
\label{alg:behavior}
\end{algorithm}
On top of all sequences presented in Algorithm~\ref{alg:behavior}, the system incorporates a STOP behavior which directly lands the platform on the ground, using either RC switch or a software safety command from the task supervisor. The pre-flight checks include bringup of hardware and all software COMPRA stack. The Breadcrumb Waypoint List $\mathbb{BW}$ is populated with exploration waypoints $wp^{\mathcal{B}}$ at a user desired sampling rate.  On the return path, the visited breadcrumb waypoints within a radius are removed from $\mathbb{BW}$. $T^{current}_{mission}$ and $T^{reference}_{mission}$ refer to current mission time and overall mission time respectively. 

For the return of the \gls{mav} to the mission's starting point, previously visited points are considered in $\mathbb{BW}$. In this case the \gls{mav}, navigates through already mapped areas, which allows relocalization or loop closure attempts to happen, processing localization drifts occurred during the navigation.

\section{Experimental Results} \label{sec:results}
\subsection{System Overview} \label{sec:setup}
\begin{figure}[!htbp]
    \centering
\includegraphics[width=0.85\columnwidth]{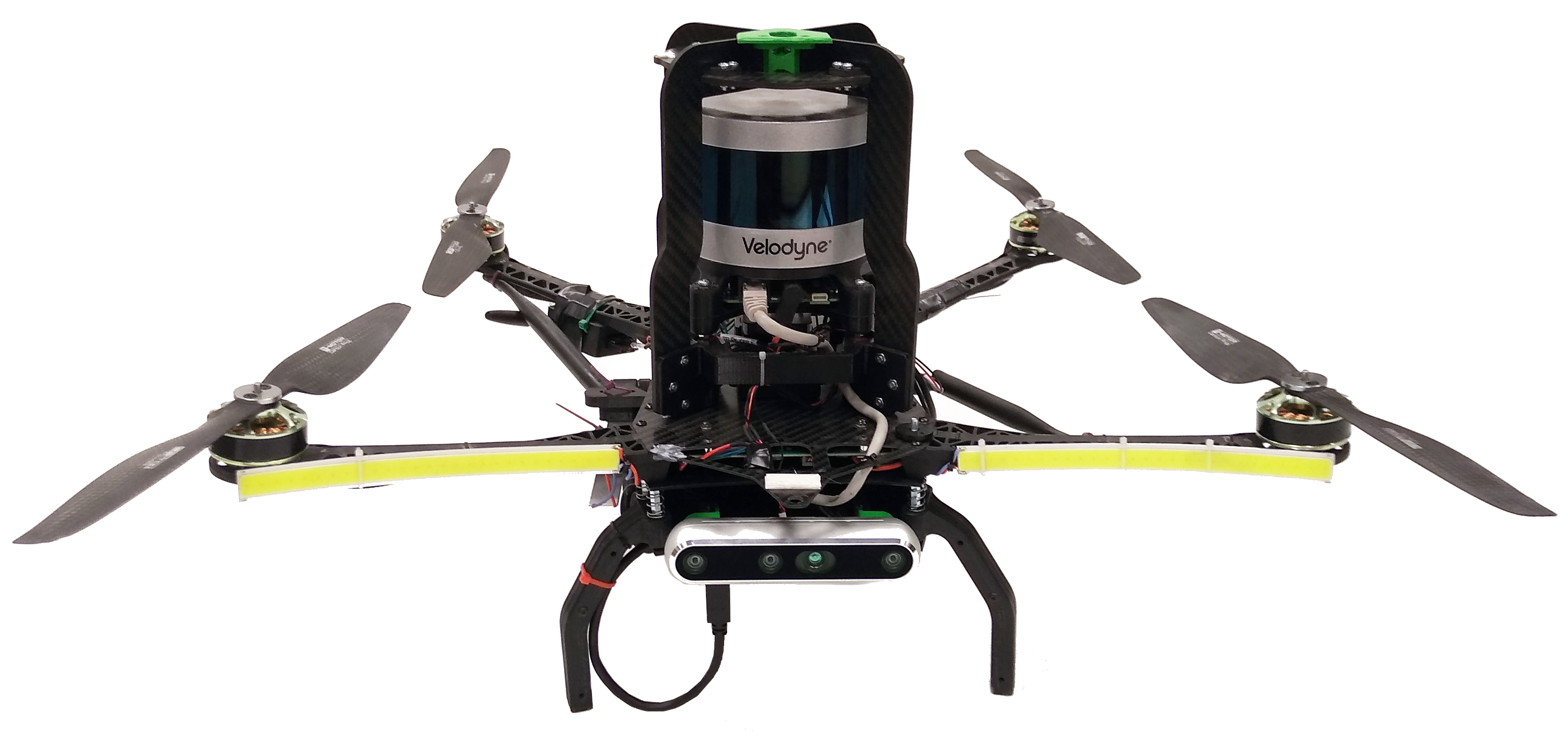}
    \caption{The aerial system utilized in the SubT field trials for evaluating the COMPRA framework.}
    \label{fig:shafter_platform}
\end{figure}

The experimental system, used for the verification of the COMPRA framework, consists of a custom built quadrotor platform, as shown in Figure~\ref{fig:shafter_platform}. The platform's mass is approximately $\unit[3.5]{kg}$, and the maximum size diameter is $\unit[0.85]{m}$, with processing payload the Intel NUC - NUC10i5FNKPA processor as well as the Intel Neural Compute Stick 2. The 3D LiDAR Velodyne VLP16 PuckLite provides 3D pointclouds at $\unit[10]{Hz}$ with $\unit[360]{^{\circ}}$ Horizontal FOV and $\unit[30]{^{\circ}}$ Vertical FOV within the range of $\unit[100]{m}$. The LiDAR measurements are used in the obstacle avoidance module, as well as in the localization module, combined with Inertial measurements. The single beam LiDAR LiDAR-Lite v3 is mounted facing towards the ground and provides relative altitude measurements. Moreover, the platform carries the RGB-D camera Intel Realsense D455 that provides RGB frames at $\unit[30]{FPS}$ with $\unit[90]{^{\circ}}$ Horizontal FOV and $\unit[65]{^{\circ}}$ Vertical FOV, as well as depth images at $\unit[30]{FPS}$ with $\unit[86]{^{\circ}}$ Horizontal FOV and $\unit[57]{^{\circ}}$ Vertical FOV, within the range of $\unit[6]{m}$, used in the object detection and heading regulation modules. The \gls{mav} is equipped with the low-level flight controller PixHawk 2.1 Black Cube, which provides the IMU measurements. Finally, the \gls{mav} is equipped with two $\unit[10]{W}$ LED light bars in the front arms for additional illumination and one $\unit[10]{W}$ LED light bars looking downwards. The COMPRA software stack has been evaluated in the Ubuntu 18.04 Operating System using the Robot Operating System (ROS)~Melodic version, while the modules have been implemented either C++, Rust, or Python.
\subsection{Field Validation Sites} \label{sec:experiments}
The proposed autonomy stack was evaluated in a variety of subterranean operating conditions (narrow, wide, curving, inclined, dusty, obstacles) at a subterranean tunnel complex located in Lule{\aa}, Sweden with parts completely lacking natural or external illumination, in the presence of other corrupting magnetic fields, while small dust particles were floating around the platform. We also include experiments from two visits to real mining environments at the Epiroc Test Mine in Örebro, Sweden, as well as at the Callio Pyhäsalmi Mine, Pyhäjärvi, Finland.

\subsection{Complete Missions}
This section will present the COMPRA-enabled \gls{mav} in full mission scenarios. In order for the reader to see the real-time behavior of the navigation and object detection pipeline we strongly suggest the reader to watch a compilation of full experiment mission runs at \url{https://www.youtube.com/watch?v=xHmeX7a8A3g}. In the following scenarios, a portable calibration gate consisted of a bundle of four AprilTags, placed in front of the vehicle, is used to initialize the global coordinate frame in which the objects of interest will be localized. All missions parts are handled completely autonomously, from initialization, arming etc. to landing after mission completion. The visualization of various mission components have gone through no post-processing and are available to the operator stationed at the mission start location as soon as the mission has been completed, which of course is a key aspect of a realistic \gls{sar} mission. A safety pilot monitors the MAV but does not interact with it in any way. 

\begin{figure}[!htbp]
    \centering
\includegraphics[width=0.9\columnwidth]{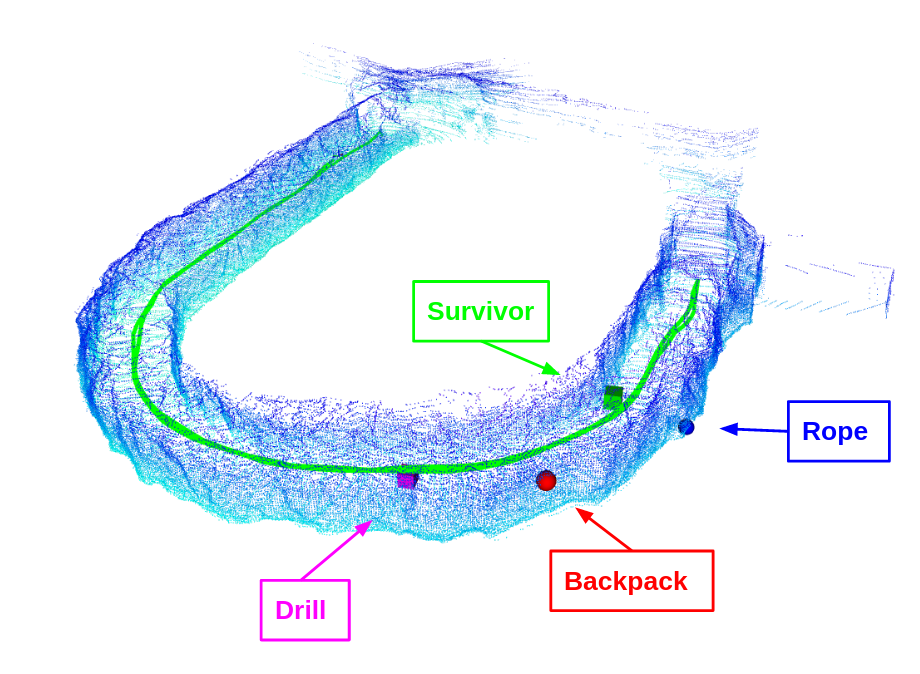}
    \caption{Visualization of complete exploration mission in a curving part of a longer tunnel, where artifacts have been placed along the tunnel. The MAV is exploring at around \unit[1]{m/s}.}
    \label{fig:rondell_mission}
\end{figure}

\begin{figure}[htbp]
    \centering
\includegraphics[width=0.9\columnwidth]{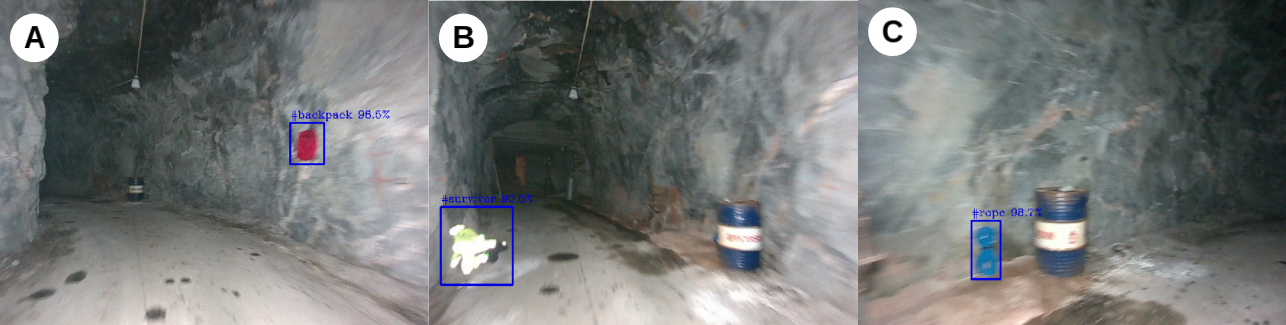}
    \caption{Artifact detection hits for mission in Figure \ref{fig:rondell_mission}}
    \label{fig:rondell_snaps}
\end{figure}

\begin{figure}[!htbp]
    \centering
\includegraphics[width=\columnwidth]{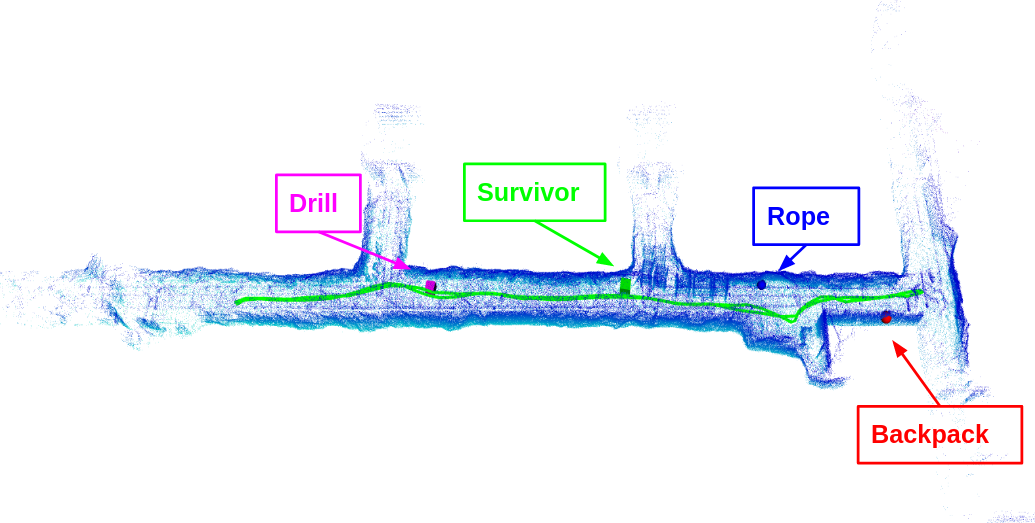}
    \caption{Visualization of mission in a tunnel with junctions and obstacles.}
    \label{fig:retreat_mission}
\end{figure}

\begin{figure}[htbp]
    \centering
\includegraphics[width=0.9\columnwidth]{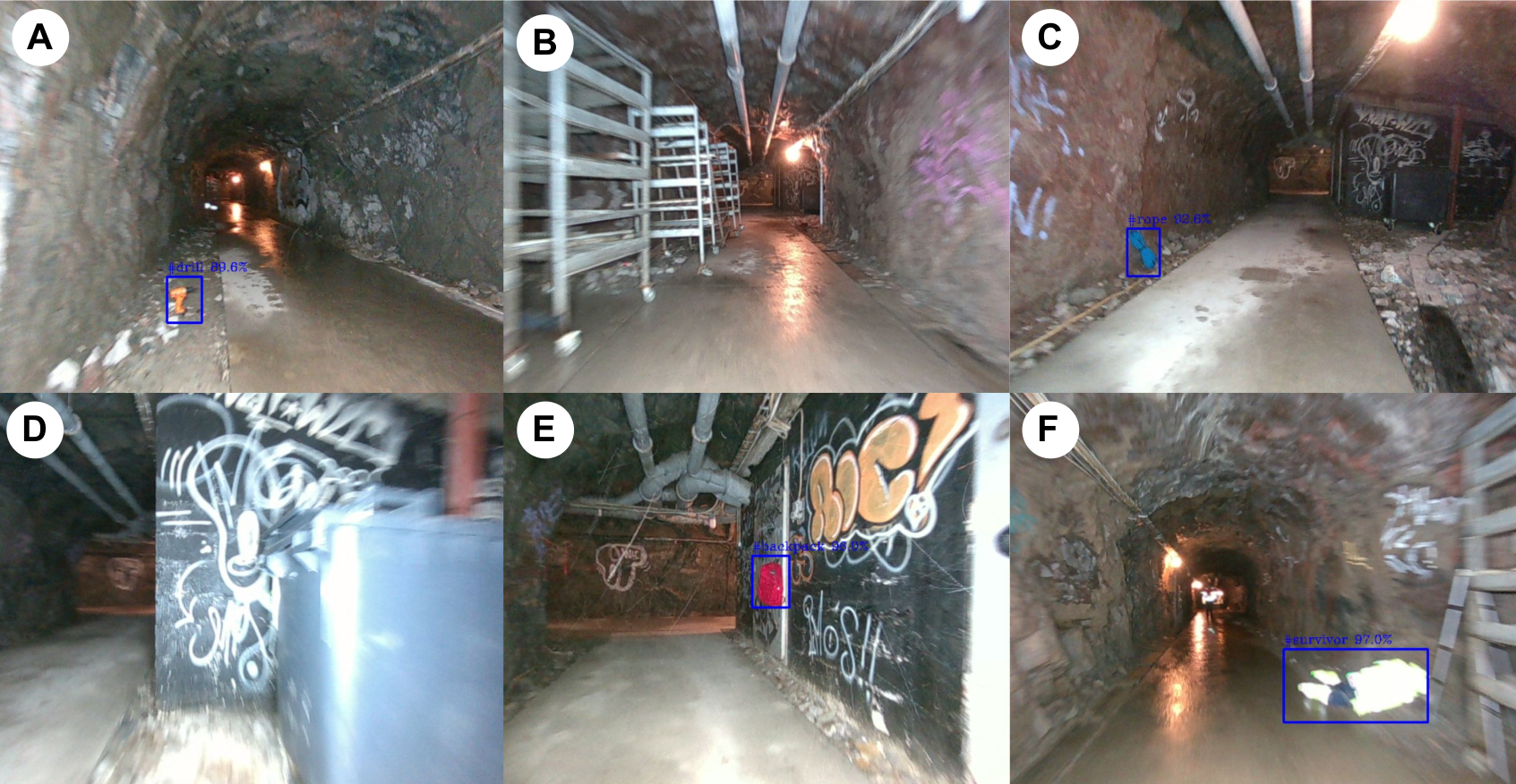}
    \caption{Artifact detection hits for mission in Figure \ref{fig:retreat_mission} (a,c,e,f), as well as obstacle avoidance scenarios (b,d), where (b) shows the narrower area of the tunnel.}
    \label{fig:retreat_snaps}
\end{figure}

\begin{figure}[!htbp]
    \centering
\includegraphics[width=\columnwidth]{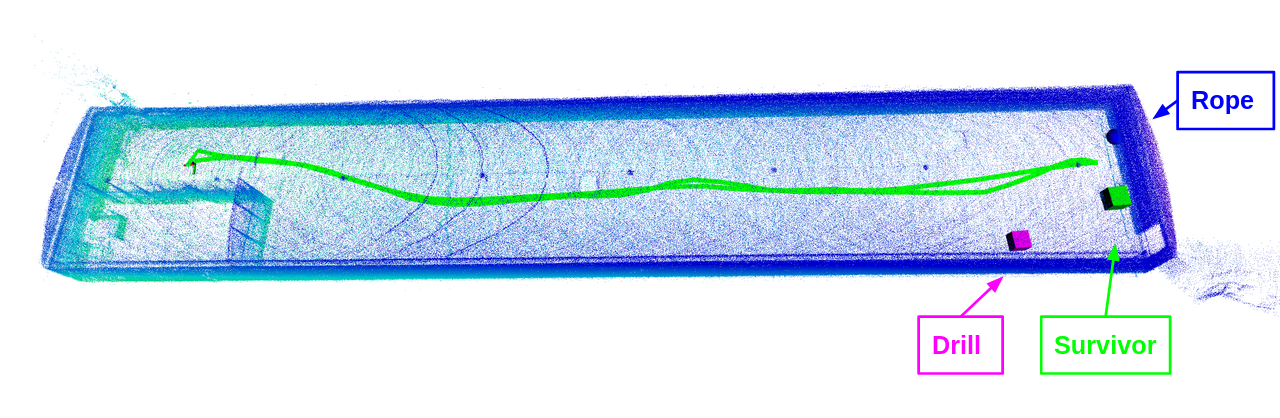}
    \caption{Visualization of mission in a larger void area $\sim\unit[45\times8\times4]{m^3}$.}
    \label{fig:void_mission}
\end{figure}

\begin{figure}[htbp]
    \centering
\includegraphics[width=0.9\columnwidth]{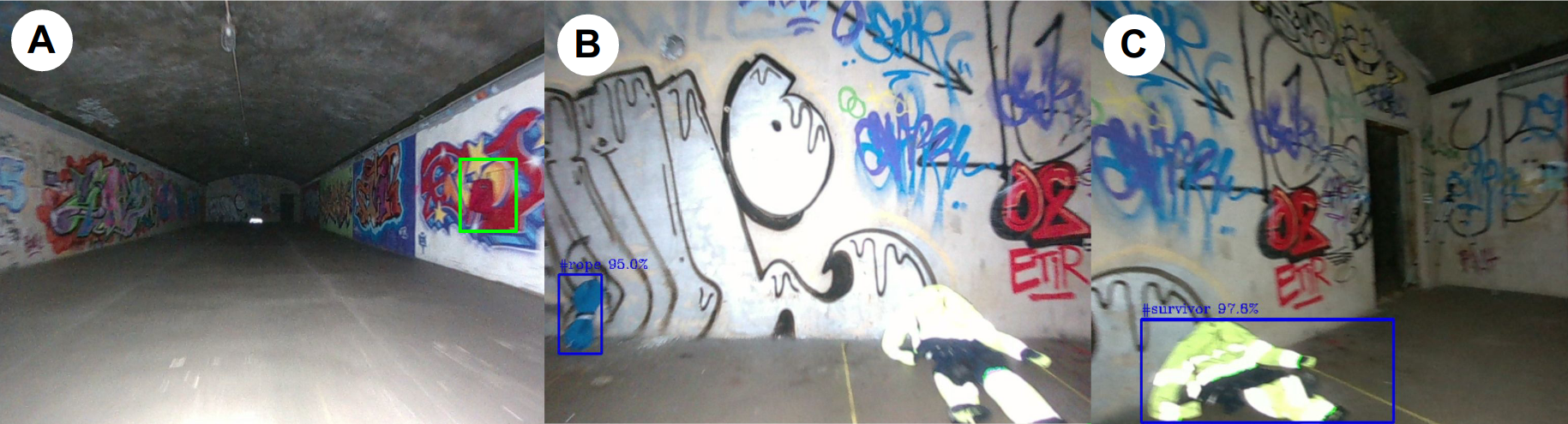}
    \caption{Graffiti-rich area to challenge object detection. Missed backpack hit (a), and artifact detection hits for mission in Figure \ref{fig:void_mission} (b,c).}
    \label{fig:void_snaps}
\end{figure}

\begin{figure}[!htbp]
    \centering
\includegraphics[width=\columnwidth]{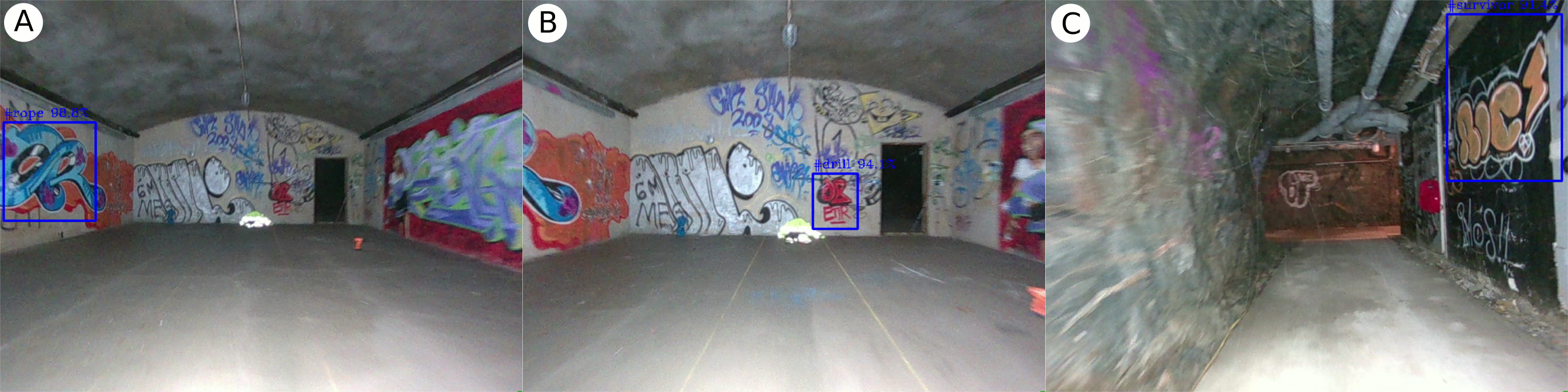}
    \caption{Visualization of rejected false positive object detection in areas with textured walls at the void and the tunnel environments. The visualization includes the classification confidence as well as the classified class.}
    \label{fig:detection_false_positives}
\end{figure}

\begin{figure}[!htbp]
    \centering
\includegraphics[width=\columnwidth]{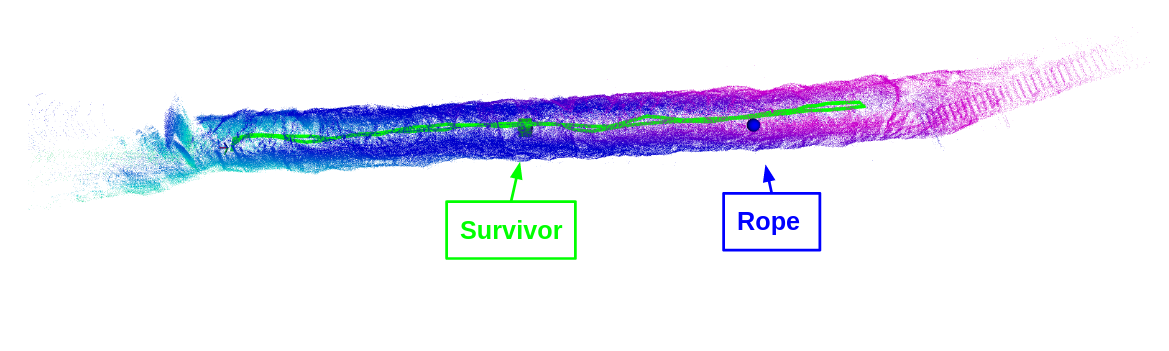}
    \caption{Visualization of mission in a narrow as well as inclined cave-like tunnel.}
    \label{fig:cave_mission}
\end{figure}

\begin{figure}[htbp]
    \centering
\includegraphics[width=0.9\columnwidth]{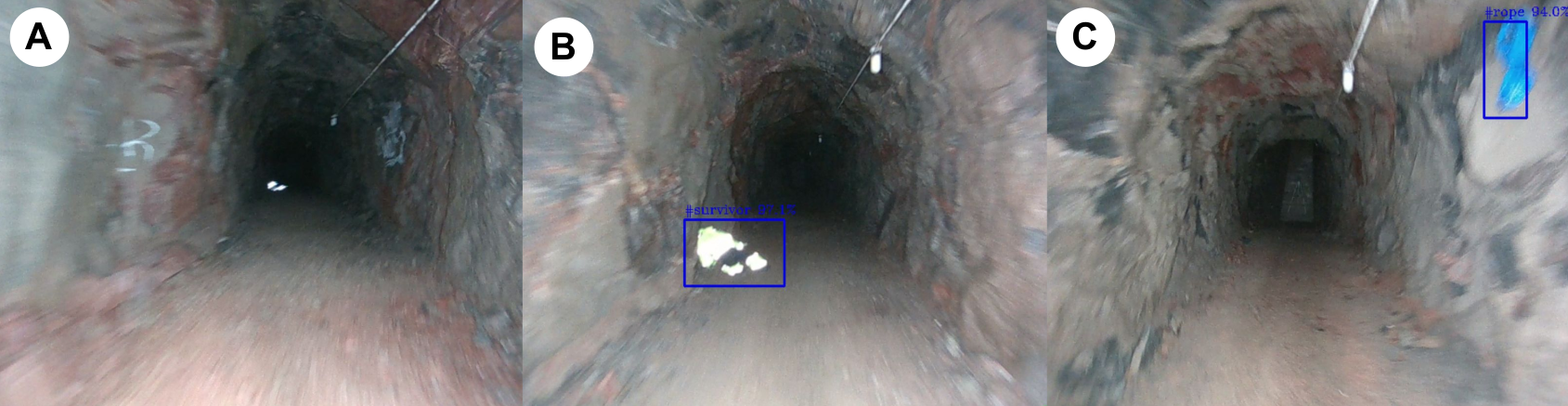}
    \caption{Narrow and inclined tunnel area (a), and artifact detection hits for mission in Figure \ref{fig:cave_mission} (b,c).}
    \label{fig:cave_snaps}
\end{figure}

\begin{figure}[!htbp]
    \centering
\includegraphics[width=\columnwidth]{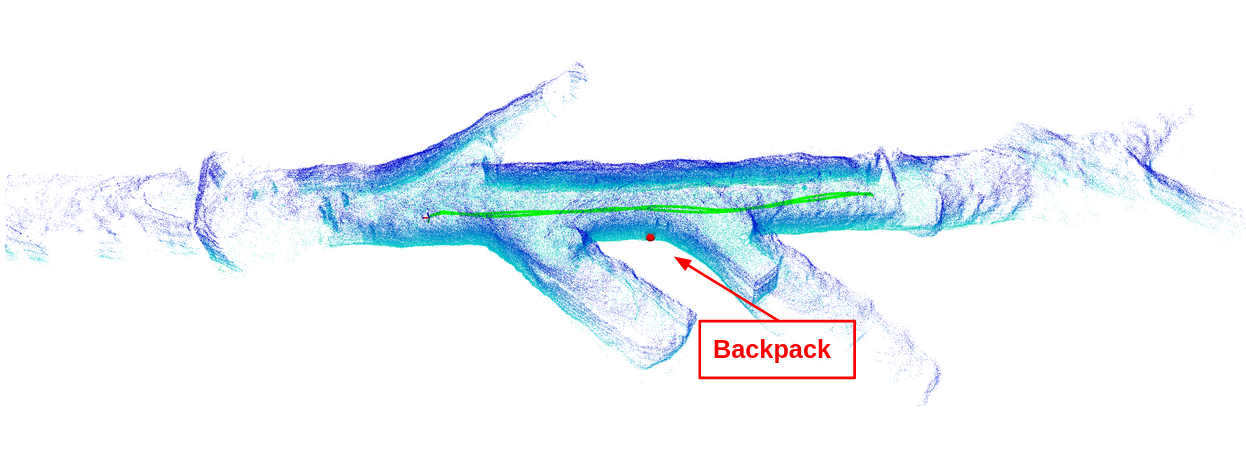}
    \caption{Visualization of shorter mission in the Callio Pyhäsalmi Mine, Pyhäjärvi, Finland.}
    \label{fig:finland_mission}
\end{figure}

\begin{figure}[htbp]
    \centering
\includegraphics[width=0.9\columnwidth]{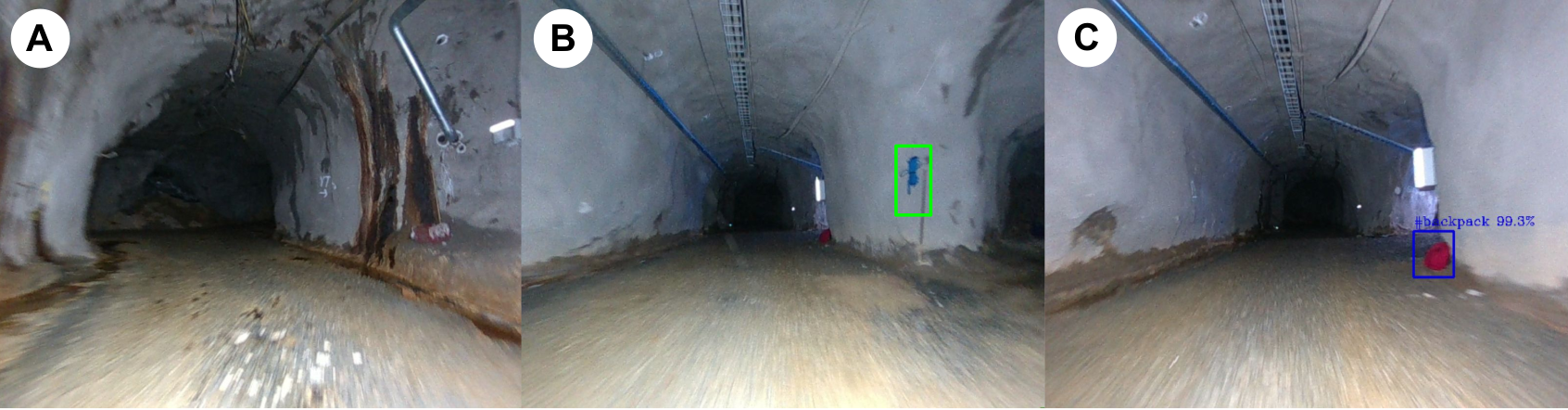}
    \caption{Real mining tunnel environment from Figure \ref{fig:finland_mission} (a), missed artifact (rope) (b) and detected backpack (c).}
    \label{fig:finland_snaps}
\end{figure}

\begin{figure}[!htbp]
    \centering
\includegraphics[width=\columnwidth]{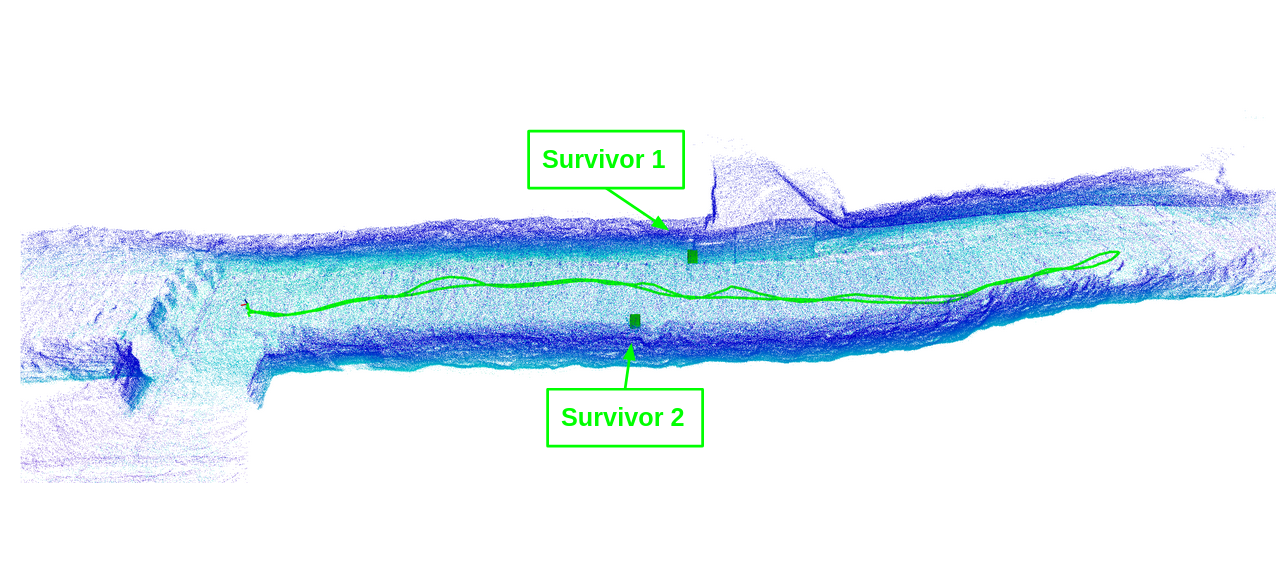}
    \caption{Visualization of mission in the Epiroc test mine, Örebro, Sweden. Wider realistic mining tunnels for large mining machines. Explored area $\sim\unit[100\times12\times5]{m^3}$.}
    \label{fig:epiroc_mission}
\end{figure}

\begin{figure}[htbp]
    \centering
\includegraphics[width=0.9\columnwidth]{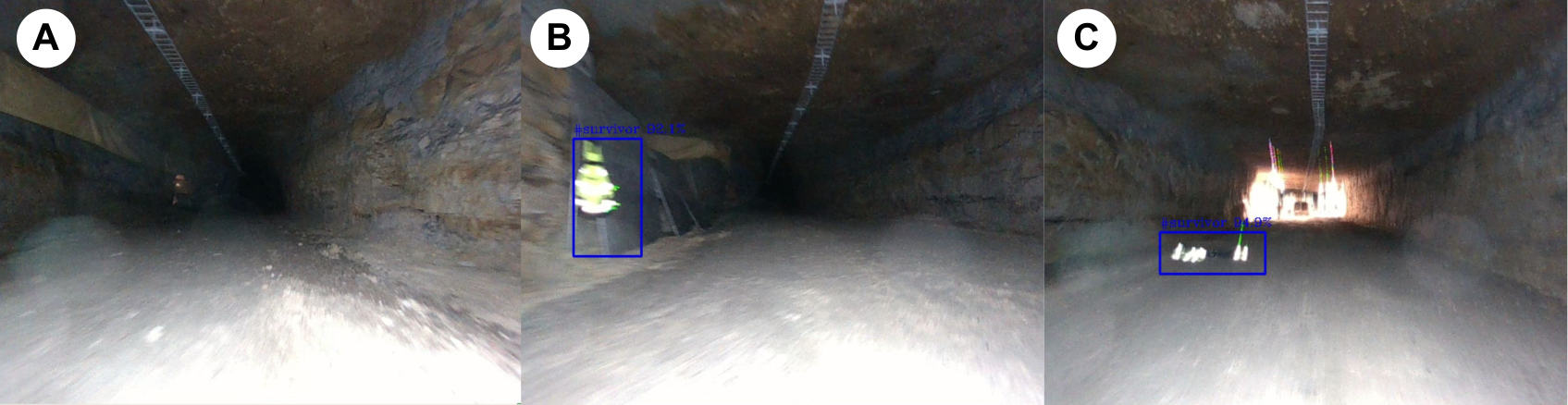}
    \caption{Wide mining tunnel environment (a), artifact detection hits during mission in Figure \ref{fig:epiroc_mission} (b,c)}
    \label{fig:epiroc_snaps}
\end{figure}

\begin{figure}[!htbp]
    \centering
\includegraphics[width=0.7\columnwidth]{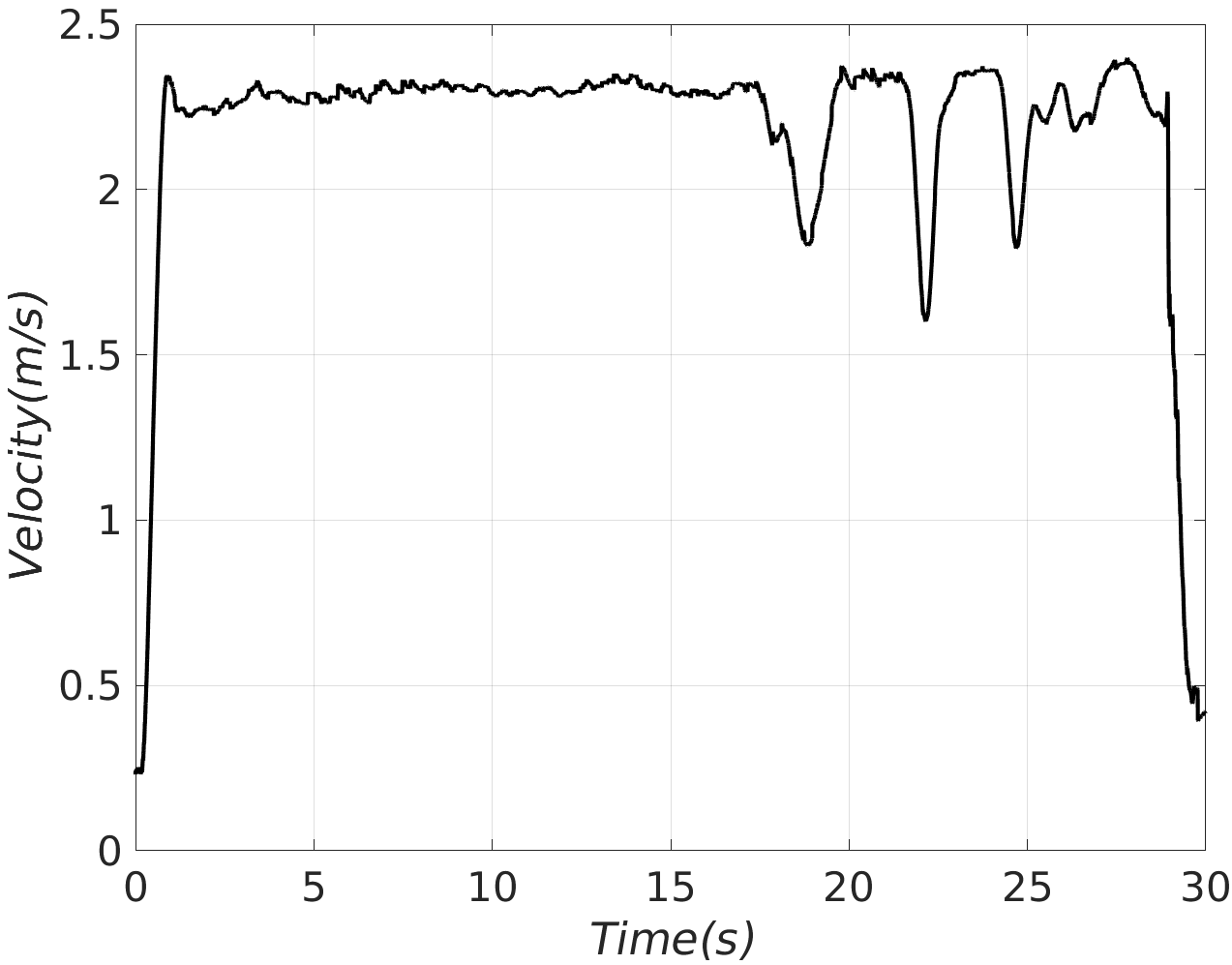}
    \caption{Magnitude of velocity for a shorter high-speed mission in the same curving tunnel area seen in Figure \ref{fig:rondell_mission}.}
    \label{fig:fast_mission}
\end{figure}

The Figures \ref{fig:rondell_mission}-\ref{fig:epiroc_snaps} display six field missions targeting different environments where Figures \ref{fig:finland_mission}-\ref{fig:epiroc_snaps} are from the real mines. We display generated pointcloud maps of the various environments where we deployed COMPRA with located artifacts (markers) and the exploration path of the MAV (green line) as well as snapshot images from the onboard camera stream showing the artifact detection hits (or misses) as well as other descriptive moments during the missions. The COMPRA-enabled MAV is set to explore areas with a speed of around $\unit[0.9 - 1]{m/s}$ in slightly wider tunnels ($\sim \unit[3.5]{m}$) such as in Figure \ref{fig:rondell_mission} or in the void-like area of Figure \ref{fig:void_mission}. In the narrower environments, such as the inclined tunnel in Figure \ref{fig:cave_mission} with a width of around $\sim\unit[1.8]{m}$ (but momentarily narrower) or in the narrowest area of the mission in Figure \ref{fig:retreat_mission} shown in Figure \ref{fig:retreat_snaps}b, the speed is reduced (but still averaging around $\unit[0.8]{m/s}$) due to the adaptive weights in the presence of obstacles. The main limiting factor for navigation speed was the object detector, where performance was reduced with an increase in navigation speed. We performed an extra shorter experiment without artifacts in the curving tunnel area (the area in Figure \ref{fig:rondell_mission}) trying to push the navigation speed. We could easily reach an average of $\unit[2.3]{m/s}$ after the initial acceleration, and the velocity (magnitude) from that mission can be seen in Figure \ref{fig:fast_mission}. In the COMPRA framework, the mission velocity is not constrained by computation time, map update rates, or similar navigation-related limitations due to its purely reactive nature, and as such to increase the velocity we simply increase $Q_{max}, Q_{min}$ for a more aggressive following of waypoints. As seen, the velocity can be kept relatively consistent throughout the mission as there are no "transition points" from one trajectory to the next since the waypoint is continually updated by the reactive exploration scheme. The dips seen in Figure \ref{fig:fast_mission} simply represent instants where the adaptive weights scheme reduced the speed due to the proximity of an obstacle, and the velocity never dropped below $\unit[1.6]{m/s}$. The smoothness of the navigation even at high velocities in constrained environments (curving mining tunnel) is a major outcome of the COMPRA framework. 

In general, the exploration and navigation behavior of the \gls{mav} was very smooth and consistent for all evaluation scenarios and provided quick, efficient, and safe tunnel navigation by keeping in the middle of the tunnel and maintaining a safe distance from walls, floor, and ceiling. This is a major outcome of the potential field formulation that keeps the \gls{mav} at a safe distance, but without any jerky or oscillatory maneuvering which can be detrimental to state estimation and object detection modules while also slowing down the exploration due to unnecessary maneuvering. The \gls{dphr} technique also proved very consistent in maintaining a fast exploration pace in the evaluated subterranean environments by its simple motion directive of aligning the \gls{mav} body \textit{x}-axis with the open areas, and keeping a continuous forward exploration.

There was one moment of risky behavior highlighted in Figure \ref{fig:retreat_snaps}d, but the \gls{apf} again provided a smooth and stable avoidance maneuver that maintained a minimum of $\unit[0.54]{m}$ distance to the obstacle. We also want to highlight the area shown in \ref{fig:retreat_snaps}b in terms of obstacle avoidance. The scaffolding-type constructions of thin steel beams are hard to detect, but as the purely reactive \gls{apf} can directly use the raw LiDAR pointcloud data for avoidance the \gls{mav} safely navigates through the risky area. Experiments performed in very wide areas, in Figures \ref{fig:void_mission} and \ref{fig:epiroc_mission}, also show that even without the \gls{apf} constantly pushing the \gls{mav} to the center of the tunnel, the exploration behavior using only the \gls{dphr} is still sufficient to follow the tunnel with minimal swaying and unnecessary movement. We should note that in these wider areas, it is only thanks to the long range of the Velodyne Puck Lite that the reactive  exploration method is sufficient for coverage of the area.

In general, the COMPRA object localizer pipeline was able to report the majority of the visited and observed artifact locations identifying 15 out of 17 artifacts in the reported experimental runs. The precision of the localization of objects was not possible to quantify during the performed experimental trials, while highly depends on the robot state estimator and the depth image analysis precision.

Nevertheless, the deployment of such detectors in realistic environments poses additional challenges, where in few locations at the void and tunnel environments were among the most challenging for the trained object detector, due to either false negative reports, shown in Figures~\ref{fig:void_snaps}(a,b) and~\ref{fig:finland_snaps}(b) or multiple false positive reports. The merit of temporally accumulating multiple observations  counteracted for the occasional false negative instances of the detector, depicted in Figure~\ref{fig:void_snaps}(b,c).

The most frequent false positive detection was observed in areas close to the textured wall, where the graffiti shape and color was similar to the target objects. In this case the size check addition within the object localization architecture allowed to discard falsely reported artifacts for localization. Figure~\ref{fig:detection_false_positives} depicts the false positive examples that COMPRA object localization pipeline recognised as outliers, and rejected them.

\subsection{Comparison Result}
\begin{figure}[htbp]
    \centering
\includegraphics[width=\columnwidth]{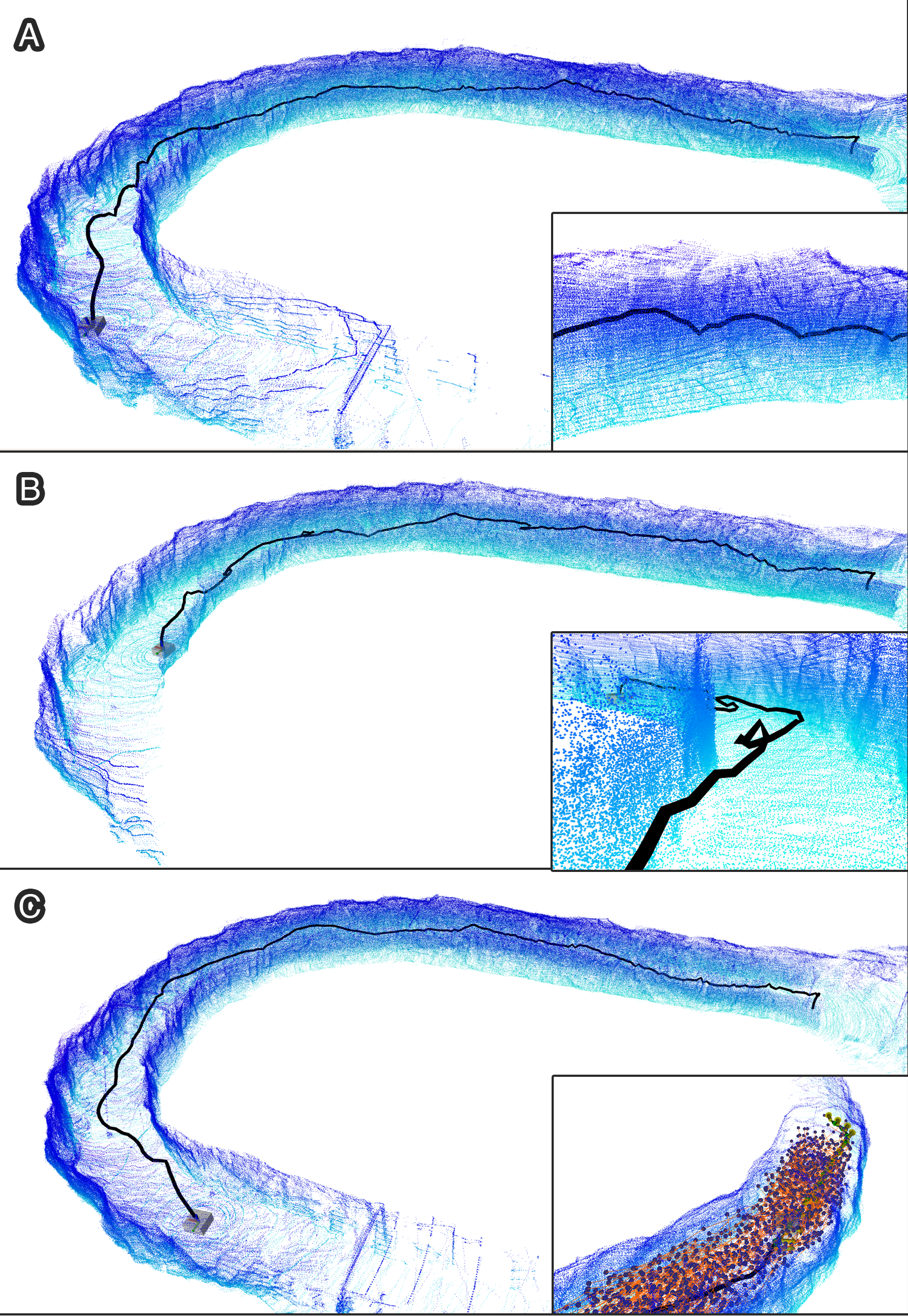}
    \caption{Resulting exploration paths using the \gls{mbp} in the curving tunnel environment, for three exploration runs.}
    \label{fig:mbplanner}
\end{figure}

\begin{figure}[htbp]
    \centering
\includegraphics[width=\columnwidth]{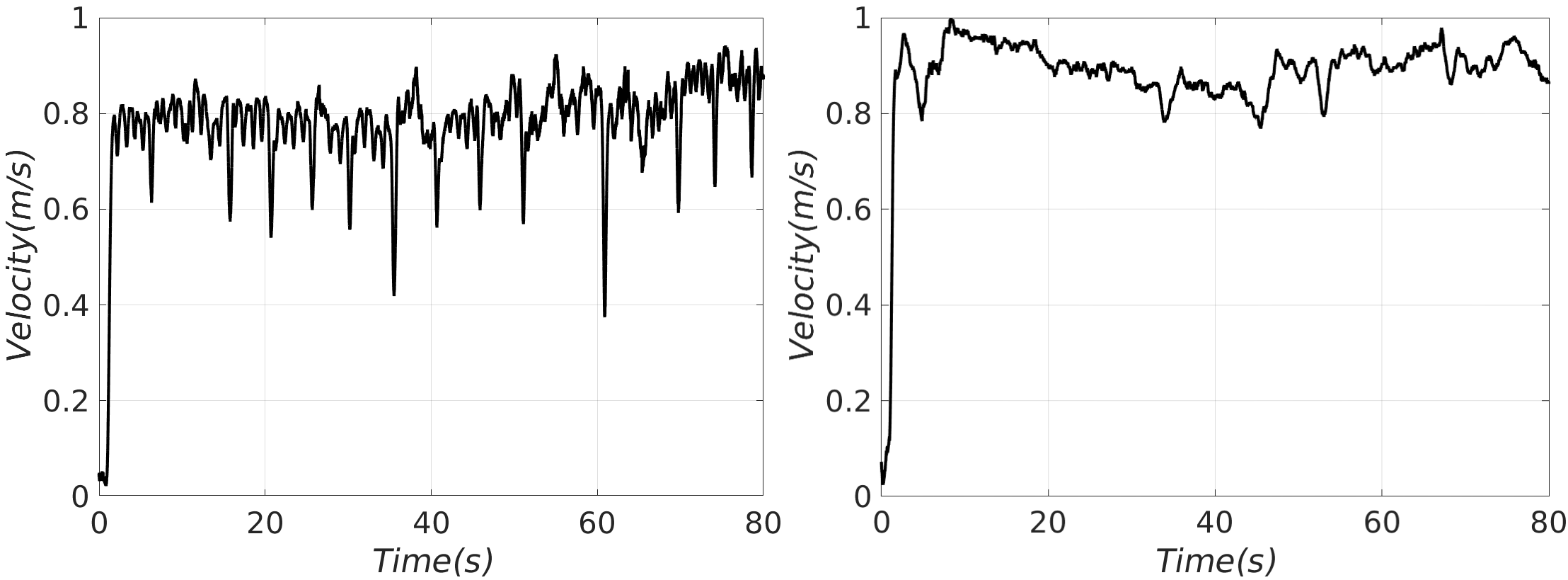}
    \caption{Velocity magnitude from \gls{mbp} experiment shown in Figure \ref{fig:mbplanner}a (left) and COMPRA run in the same area (right) with approximately the same average velocity.}
    \label{fig:mbp_vel}
\end{figure}

Performing comparisons with another complete \gls{sar} framework is very difficult due to varying sensor suites and \gls{mav} design, not all of the software being open-source, as well as tuning based on the selection of experiment site. Instead we offer a comparison focusing specifically on the ability of the COMPRA framework to efficiently and quickly navigate the tunnel environment, using the \gls{mbp} \cite{dharmadhikari2020motion}, specifically designed for agile and fast subterranean navigation in tightly constrained environments. The \gls{mbp} represents, in the author's opinion, the absolute state-of-the-art for general exploratory local path planning. The \gls{mbp} is based on the occupancy mapper Voxblox \cite{oleynikova2017voxblox} and uses an information-gain maximizing formulation combined with a clever actuation (acceleration) sampling method based on tree expansion that generates trajectories that promote agile and fast locomotion for a general solution to the combined exploration and path planning problem. We still have to mention that we are relying on our own MAV which is considerably larger than in the comparison work, a different source of state-estimation, and in order to match with our framework also a different trajectory following controller, so the comparison is still not identical to the related work. As such we will also focus on numbers provided in \cite{dharmadhikari2020motion}, while comparison missions using the \gls{mbp} are meant mainly to emphasize the difference in navigation strategy and showcase instances of how those differences manifest in the real-life navigation behavior.
Our selection of tuning parameters were: representing the UAV as a box of size $\unit[1\times1\times0.5]{m}$ (for safety distances), a desired velocity of $\unit[1]{m/s}$, and a path step length of $\unit[0.7]{m}$. We deployed the \gls{mbp} in the curving longer tunnel shown in Figure \ref{fig:rondell_mission} for three experiment runs. The generated pointcloud maps and exploration paths are shown in Figure \ref{fig:mbplanner}, while Figure \ref{fig:mbplanner}c also shows the expanded local search graph. While the \gls{mbp} paths efficiently maximize information gain while largely maintaining a forward locomotion and efficiently avoiding collisions, there are some moments or instances of the same difficulties that most map-based local exploration/path planning frameworks face, that being: unnecessary (not forward) movement due to either the random nature of generated graphs or from trying to get some specific frontier point into sensor field-of-view (ex. last part of the path Figure \ref{fig:mbplanner}a), rough transitions between trajectories (ex. in \ref{fig:mbplanner}b), and in general not as smooth navigation as the missions using the COMPRA framework, exemplified by the velocity profile comparison from the same area in Figure \ref{fig:mbp_vel}. But, notably, the mission run in Figure \ref{fig:mbplanner}c had very limited issues while using the exact same tuning. On the topic of very agile motion the related work \cite{dharmadhikari2020motion} shows experimental results for up to $\unit[1.8]{m/s}$ average exploration speed as compared to the $\unit[2.3]{m/s}$ using COMPRA in a similar tunnel environment. High-speed navigation in narrow or constrained subterranean environments is a very difficult problem, and other related works are evaluated (in real-life constrained environments) up to: $\unit[0.1]{m/s}$~\cite{mansouri2020deploying}, $\unit[0.4-0.5]{m/s}$~\cite{kratky2021autonomous}, $\unit[0.5]{m/s}$~\cite{petrlik2020robust}, ~$\unit[0.5]{m/s}$~\cite{ozaslan2017autonomous}, $\unit[0.75]{m/s}$~\cite{dang2019graph}, as compared to the $\unit[0.9-1]{m/s}$ full mission speed and the $\unit[2.3]{m/s}$ navigation evaluation speed of COMPRA. 
Additionally, COMPRA could maintain a relatively high velocity when passing through the narrower spaces and required no special tuning to do so but, again, comparisons are hard due to different MAV sizes, lack of propeller guards on our \gls{mav} promoting a larger critical safety distance, different environments etc.

The \gls{mbp} represents a more general framework for exploration than the compact algorithmic design of COMPRA and would likely outperform it in many environments, but when we can utilize the tunnel morphology to generate the desired behavior for rapid and directed tunnel navigation and exploration, the proposed fully reactive scheme has some advantages, while also being decoupled from any mapping software.

\subsection{Discussion \& Lessons Learned}\label{sec:limitations}
During the field tests we had to address issues related to state estimation, heading regulation, artifact detection and flight time of the platform. More specifically, the underground areas that the \gls{mav} explored was quite challenging with respect to the geometry and self-similar shape in terms of the LiDAR-Inertial method. Overall, we tuned the LiDAR-Inertial method to operate in underground environments, while we aim to enhance the methodology to support large-scale and long-term operations incorporating sensor fusion in the pipeline. A challenge in terms of obstacle avoidance is achieving the desired flight behavior both in obstacle free areas where the aim is moving as quickly as possible, and in obstacle rich areas, where stable avoidance behavior should be prioritized over speed. While we include an initial method on adapting weights on position reference tracking to speed up or slow down the exploration depending on the proximity of obstacles, further adaptive tuning on NMPC parameters as well as \gls{apf} gains and safety distances depending on the environment and current maneuvering of the platform should be investigated. This would allow even more adaptive behavior in different circumstances which increases the range of possible deployment scenarios for the reactive framework (for example, going into even narrower tunnels without conflicts with desired safety distances).
Additionally, developing a framework of mission planning that utilizes the fully reactive and fast exploration of COMPRA in combination with higher-level directives to handle junctions and the general question of completely exploring larger subterranean systems, is a key future work. This framework could use the combination of \gls{apf} and \gls{dphr} for consistent and efficient "junction-to-junction" navigation, e.g. still decoupling the local navigation behavior from occupancy mapping and path planning but integrating map-based junction detection and mission behavior.
 Finally, the flight time of the aerial platform is critical for the mission and game changer when designing autonomy frameworks. COMPRA assumes platforms with limited flight time, aiming for quick deployment as well as fast navigation and scouting behaviour of unexplored areas, instead of detailed area coverage.  
\section{Conclusions} \label{sec:conclusion}
This work presented COMPRA navigation framework for \glspl{mav}, targeting \gls{sar} operations in subterranean environments. The COMPRA framework enables the fast deployment of a fully autonomous \gls{mav} to navigate along previously unknown tunnel areas. The proposed autonomy architecture combines 3D LiDAR-inertial state estimation, position based \gls{nmpc} control, potential field based obstacle avoidance, reactive exploration based on heading regulation, as well as object detection, localization and mission behaviour. COMPRA's main aim is to keep the overall system compact, reactive, resilient, independant on occupancy mapping, and with low complexity as a baseline solution for a complete SubT exploration and object localization mission, while it can be added upon in a straight-forward way by further developments of higher-level mission planner modules. Multiple field trials in a variety of subterranean environments successfully demonstrate the performance and efficacy of the method towards real-life applications in realistic subterranean tunnels and voids, showing fast and consistent navigation behavior in conjunction with object detection and mission behavior, despite the challenging environments.

\section*{Declarations}

\textbf{Funding}: This work has been partially funded by the European Unions Horizon 2020 Research and Innovation Programme under the Grant Agreement No. 869379 illuMINEation.\\
\textbf{Conflict of interest}: The authors have no conflicts of interest with any related parties.\\
\textbf{Competing Interests}: Not applicable\\
\textbf{Availability of data and material:} ROSbags of sensor and submodule output data from field experiments can be made available at the suggestion of the reviewers and editors. \\
\textbf{Authors' Contributions}: Björn Lindqvist and Christoforos Kanellakis: Development, implementation, system integration and field work, relating to all presented submodules and developments, main manuscript contributors. Sina Sharif Mansouri: Software development and field expertise. Ali-akbar Agha-mohammadi: Advisory, development lead for Team CoSTAR in DARPA SubT Challenge. George Nikolakopoulos: Advisory, manuscript contributions, head of Luleå University of Technology Robotics\&AI Team. All authors have read and approved the manuscript.\\
\textbf{Ethics approval}: Not applicable.\\
\textbf{Consent to Participate}: Not applicable.\\
\textbf{Consent to Publish}: All authors comply with the consent to publish.\\

\begin{acknowledgements}
This work has been partially funded in part by the European Unions Horizon 2020 Research and Innovation Programme under the Grant Agreements No. 869379 illuMINEation, No.101003591 NEXGEN-SIMS and in part by the Interreg Nord Programme ROBOSOL NYPS 20202891.
\end{acknowledgements}

\bibliography{mybib}

\begin{thebibliography}{10}
\providecommand{\url}[1]{{#1}}
\providecommand{\urlprefix}{URL }
\expandafter\ifx\csname urlstyle\endcsname\relax
  \providecommand{\doi}[1]{DOI \discretionary{}{}{}#1}\else
  \providecommand{\doi}{DOI \discretionary{}{}{}\begingroup
  \urlstyle{rm}\Url}\fi

\bibitem{kalantari2020drivocopter}
A.~Kalantari, T.~Touma, L.~Kim, R.~Jitosho, K.~Strickland, B.T. Lopez, A.A.
  Agha-Mohammadi, in \emph{2020 IEEE Aerospace Conference} (IEEE, 2020), pp.
  1--10

\bibitem{thakur2018nuclear}
D.~Thakur, G.~Loianno, W.~Liu, V.~Kumar, in \emph{International Symposium on
  Experimental Robotics} (Springer, 2018), pp. 191--200

\bibitem{mansouri2018cooperative}
S.S. Mansouri, C.~Kanellakis, E.~Fresk, D.~Kominiak, G.~Nikolakopoulos, Control
  Engineering Practice \textbf{74}, 118 (2018)

\bibitem{rogers2017distributed}
J.G. Rogers, R.E. Sherrill, A.~Schang, S.L. Meadows, E.P. Cox, B.~Byrne, D.G.
  Baran, J.W. Curtis, K.M. Brink, in \emph{Ground/Air Multisensor
  Interoperability, Integration, and Networking for Persistent ISR VIII}, vol.
  10190 (International Society for Optics and Photonics, 2017), vol. 10190, p.
  1019017

\bibitem{subt}
DARPA.
\newblock Subterranean challenge {(SubT)}.
\newblock \urlprefix\url{https://www.subtchallenge.com/}

\bibitem{nebula}
{TEAM COSTAR}.
\newblock {NeBula} autonomy.
\newblock \urlprefix\url{https://costar.jpl.nasa.gov/}

\bibitem{agha2021nebula}
A.~Agha, K.~Otsu, B.~Morrell, D.D. Fan, R.~Thakker, A.~Santamaria-Navarro, S.K.
  Kim, A.~Bouman, X.~Lei, J.~Edlund, et~al., arXiv preprint arXiv:2103.11470
  (2021)

\bibitem{palieri2020locus}
M.~Palieri, B.~Morrell, A.~Thakur, K.~Ebadi, J.~Nash, A.~Chatterjee,
  C.~Kanellakis, L.~Carlone, C.~Guaragnella, A.a. Agha-Mohammadi, IEEE Robotics
  and Automation Letters \textbf{6}(2), 421 (2020)

\bibitem{kim2021plgrim}
S.K. Kim, A.~Bouman, G.~Salhotra, D.D. Fan, K.~Otsu, J.~Burdick, A.a.
  Agha-mohammadi, in \emph{Proceedings of the International Conference on
  Automated Planning and Scheduling}, vol.~31 (2021), vol.~31, pp. 652--662

\bibitem{mac2016heuristic}
T.T. Mac, C.~Copot, D.T. Tran, R.~De~Keyser, Robotics and Autonomous Systems
  \textbf{86}, 13 (2016)

\bibitem{zhao2018survey}
Y.~Zhao, Z.~Zheng, Y.~Liu, Knowledge-Based Systems \textbf{158}, 54 (2018)

\bibitem{quan2020survey}
L.~Quan, L.~Han, B.~Zhou, S.~Shen, F.~Gao, IET Cyber-systems and Robotics
  \textbf{2}(1), 14 (2020)

\bibitem{kratky2021autonomous}
V.~Kr{\'a}tk{\`y}, P.~Petr{\'a}{\v{c}}ek, T.~B{\'a}{\v{c}}a, M.~Saska, Journal
  of Field Robotics  (2021)

\bibitem{ohradzansky2021multi}
M.T. Ohradzansky, E.R. Rush, D.G. Riley, A.B. Mills, S.~Ahmad, S.~McGuire,
  H.~Biggie, K.~Harlow, M.J. Miles, E.W. Frew, et~al., arXiv preprint
  arXiv:2110.04390  (2021)

\bibitem{dang2020autonomous}
T.~Dang, F.~Mascarich, S.~Khattak, H.~Nguyen, H.~Nguyen, S.~Hirsh, R.~Reinhart,
  C.~Papachristos, K.~Alexis, in \emph{2020 IEEE Aerospace Conference} (IEEE,
  2020), pp. 1--8

\bibitem{petrlik2020robust}
M.~Petrl{\'\i}k, T.~B{\'a}{\v{c}}a, D.~He{\v{r}}t, M.~Vrba, T.~Krajn{\'\i}k,
  M.~Saska, IEEE Robotics and Automation Letters \textbf{5}(2), 2169 (2020)

\bibitem{sandino2020uav}
J.~Sandino, F.~Vanegas, F.~Maire, P.~Caccetta, C.~Sanderson, F.~Gonzalez,
  Remote Sensing \textbf{12}(20), 3386 (2020)

\bibitem{ozaslan2017autonomous}
T.~{\"O}zaslan, G.~Loianno, J.~Keller, C.J. Taylor, V.~Kumar, J.M. Wozencraft,
  T.~Hood, IEEE Robotics and Automation Letters \textbf{2}(3), 1740 (2017)

\bibitem{mansouri2020deploying}
S.S. Mansouri, C.~Kanellakis, D.~Kominiak, G.~Nikolakopoulos, Robotics and
  Autonomous Systems \textbf{126}, 103472 (2020)

\bibitem{kohlbrecher2011flexible}
S.~Kohlbrecher, O.~Von~Stryk, J.~Meyer, U.~Klingauf, in \emph{2011 IEEE
  international symposium on safety, security, and rescue robotics} (IEEE,
  2011), pp. 155--160

\bibitem{redmon2016you}
J.~Redmon, S.~Divvala, R.~Girshick, A.~Farhadi, in \emph{Proceedings of the
  IEEE conference on computer vision and pattern recognition} (2016), pp.
  779--788

\bibitem{hess2016real}
W.~Hess, D.~Kohler, H.~Rapp, D.~Andor, in \emph{2016 IEEE International
  Conference on Robotics and Automation (ICRA)} (IEEE, 2016), pp. 1271--1278

\bibitem{kamel2017model}
M.~Kamel, T.~Stastny, K.~Alexis, R.~Siegwart, in \emph{Robot operating system
  (ROS)} (Springer, 2017), pp. 3--39

\bibitem{khattak2020keyframe}
S.~Khattak, C.~Papachristos, K.~Alexis, Journal of Field Robotics
  \textbf{37}(4), 552 (2020)

\bibitem{dang2019field}
T.~Dang, F.~Mascarich, S.~Khattak, H.~Nguyen, N.~Khedekar, C.~Papachristos,
  K.~Alexis, Field and Service Robotics (FSR)  (2019)

\bibitem{dang2019graph}
T.~Dang, F.~Mascarich, S.~Khattak, C.~Papachristos, K.~Alexis, in \emph{2019
  IEEE/RSJ International Conference on Intelligent Robots and Systems (IROS)}
  (IEEE, 2019), pp. 3105--3112

\bibitem{meier2015px4}
L.~Meier, D.~Honegger, M.~Pollefeys, in \emph{2015 IEEE international
  conference on robotics and automation (ICRA)} (IEEE, 2015), pp. 6235--6240

\bibitem{lee2013nonlinear}
T.~Lee, M.~Leok, N.H. McClamroch, Asian Journal of Control \textbf{15}(2), 391
  (2013)

\bibitem{mellinger2011minimum}
D.~Mellinger, V.~Kumar, in \emph{2011 IEEE international conference on robotics
  and automation} (IEEE, 2011), pp. 2520--2525

\bibitem{kanellakis2018towards}
C.~Kanellakis, S.S. Mansouri, G.~Georgoulas, G.~Nikolakopoulos, in
  \emph{International Conference on Robotics in Alpe-Adria Danube Region}
  (Springer, 2018), pp. 173--180

\bibitem{lindqvist2020nonlinear}
B.~Lindqvist, S.S. Mansouri, A.a. Agha-mohammadi, G.~Nikolakopoulos, IEEE
  Robotics and Automation Letters \textbf{5}(4), 6001 (2020)

\bibitem{small2019aerial}
E.~Small, P.~Sopasakis, E.~Fresk, P.~Patrinos, G.~Nikolakopoulos, in \emph{2019
  18th European Control Conference (ECC)} (2019), pp. 3556--3563

\bibitem{shan2020lio}
T.~Shan, B.~Englot, D.~Meyers, W.~Wang, C.~Ratti, D.~Rus, in \emph{IEEE/RSJ
  International Conference on Intelligent Robots and Systems (IROS)} (2020)

\bibitem{karlsson2021d}
S.~Karlsson, A.~Koval, C.~Kanellakis, A.a. Agha-mohammadi, G.~Nikolakopoulos,
  arXiv preprint arXiv:2112.05563  (2021)

\bibitem{sopasakis2020open}
P.~Sopasakis, E.~Fresk, P.~Patrinos, arXiv preprint arXiv:2003.00292  (2020)

\bibitem{Hermans:IFAC:2018}
B.~Hermans, P.~Patrinos, G.~Pipeleers, IFAC-PapersOnLine \textbf{51}(20), 234
  (2018)

\bibitem{warren1989global}
C.W. Warren, in \emph{1989 IEEE International Conference on Robotics and
  Automation} (IEEE Computer Society, 1989), pp. 316--317

\bibitem{Soille_2003}
P.~Soille, \emph{Morphological Image Analysis: Principles and Applications},
  2nd edn. (Springer-Verlag, Berlin, Heidelberg, 2003)

\bibitem{openvino}
{Intel}.
\newblock {OpenVINO™ Toolkit}.
\newblock \urlprefix\url{https://docs.openvinotoolkit.org/latest/index.html}

\bibitem{bochkovskiy2020yolov4}
A.~Bochkovskiy, C.Y. Wang, H.Y.M. Liao, arXiv preprint arXiv:2004.10934  (2020)

\bibitem{dharmadhikari2020motion}
M.~Dharmadhikari, T.~Dang, L.~Solanka, J.~Loje, H.~Nguyen, N.~Khedekar,
  K.~Alexis, in \emph{2020 IEEE International Conference on Robotics and
  Automation (ICRA)} (IEEE, 2020), pp. 179--185

\bibitem{oleynikova2017voxblox}
H.~Oleynikova, Z.~Taylor, M.~Fehr, R.~Siegwart, J.~Nieto, in \emph{2017
  IEEE/RSJ International Conference on Intelligent Robots and Systems (IROS)}
  (IEEE, 2017), pp. 1366--1373

\end{thebibliography}
\end{document}